%% file: acl_latex.tex
\documentclass[11pt]{article}

\usepackage[preprint]{acl}
\usepackage{times}
\usepackage{latexsym}

\usepackage[T1]{fontenc}

\usepackage[utf8]{inputenc}

\usepackage{microtype}

\usepackage{inconsolata}

\usepackage{hyperref}       
\usepackage{url}            
\usepackage{nicefrac}       
\usepackage{colortbl}  
\usepackage{wrapfig} %

\usepackage{titlesec}
\usepackage{graphicx}
\usepackage{bm}
\usepackage{enumitem}
\usepackage{booktabs}
\usepackage{array} 
\usepackage{ragged2e}
\usepackage{multirow}
\usepackage{amsmath,amsfonts}
\usepackage{algorithm}
\usepackage{algpseudocode}
\usepackage{makecell}
\usepackage{textcomp}
\usepackage{xcolor}
\usepackage{caption}
\usepackage{subcaption}
\usepackage{amssymb}

\usepackage[most]{tcolorbox}
\definecolor{darkgrey}{rgb}{0.53,0.53,0.53}
\definecolor{mygrey}{rgb}{0.9,0.9,0.9}

\title{EduDial: Constructing a Large-scale Multi-turn \\ Teacher–Student Dialogue Corpus}

\author{
    \textbf{Shouang Wei}\textsuperscript{1}, 
    \textbf{Min Zhang}\textsuperscript{1,$\ddagger$,$\dagger$}, 
    \textbf{Xin Lin}\textsuperscript{1,$\dagger$},
    \textbf{Bo Jiang}\textsuperscript{1}, 
    \textbf{Zhongxiang Dai}\textsuperscript{2}, 
    \textbf{Kun Kuang\textsuperscript{3}} \\
    \textsuperscript{1}East China Normal University,\\
    \textsuperscript{2}The Chinese University of Hong Kong, Shenzhen,
    \textsuperscript{3}Zhejiang University\\
    \small \textsuperscript{$\ddagger$}Project leader \quad 
    \href{52285901019@stu.ecnu.edu.cn}{52285901019@stu.ecnu.edu.cn} \quad
    \textsuperscript{$\dagger$}Corresponding authors: \href{mailto:mzhang@cs.ecnu.edu.cn}{mzhang@cs.ecnu.edu.cn}, \href{xlin@cs.ecnu.edu.cn}{xlin@cs.ecnu.edu.cn}
}

\begin{document}
\maketitle


\begin{abstract}
Recently, several multi-turn dialogue benchmarks have been proposed to evaluate the conversational abilities of large language models (LLMs). As LLMs are increasingly recognized as key technology for advancing intelligent education, owing to their ability to deeply understand instructional contexts and provide personalized guidance, constructing dedicated teacher-student dialogue benchmarks has become particularly important. To this end, we present EduDial, a comprehensive multi-turn teacher-student dialogue dataset. EduDial covers 345 core knowledge points and consists of 34,250 dialogue sessions from interactions between teacher and student agents. Its design is guided by Bloom's taxonomy of educational objectives and incorporates ten questioning strategies \textemdash including situational questioning, zone of proximal development (ZPD) questioning, and metacognitive questioning \textemdash better capturing authentic classroom interactions. Furthermore, we design differentiated teaching strategies for students at different cognitive levels, providing more targeted guidance. Building on EduDial, we develop EduDial-LLM 32B via training and propose an 11-dimensional evaluation framework that systematically measures the teaching abilities of LLMs, encompassing overall teaching quality and content quality. Experiments on 17 mainstream LLMs reveal that most models struggle in student-centered teaching scenarios, whereas EduDial-LLM achieves significant gains, consistently outperforming all baselines across all metrics. Code is available at \url{https://github.com/ECNU-RAIL/EduDial-EMNLP2026}.
\end{abstract}

\input{section/introduction}

\input{section/relatedwork}

\input{section/method}

\input{section/experiment}

\input{section/conclusion}

\section*{Limitations}

Although our method achieves promising results in improving LLM-based teaching dialogues, several limitations warrant further investigation.

\begin{itemize}
    \item Our teaching framework is primarily teacher-led, advancing instruction through questioning and guidance. However, effective teaching should also encourage students to raise questions, express confusion, and challenge ideas. Future work should explore dialogue mechanisms supporting student-initiated inquiry, fostering curiosity and critical thinking.

    \item Our framework conducts teaching interactions through text-based dialogues, aligning with current large language models' core capabilities. However, human teaching often leverages multiple representational modalities, such as visualization of geometric figures, step-by-step presentation of algebraic derivations, and emotional support conveyed through vocal intonation. Future research should explore integrating visual, auditory, and other multimodal information into teaching dialogue systems to accommodate diverse learning content and student preferences.
    
   \item We evaluated the model's teaching capabilities in a controlled simulation environment where student agents respond according to predefined cognitive characteristics. While this enables systematic assessment across different student types, real classroom environments are more complex and dynamic. Future research should deploy and test EduDial in actual teaching scenarios, collecting authentic data to validate and refine the model.

\end{itemize}

\section*{Acknowledgments}
This study was funded by Guangxi Science and Technology Program (2025AB25069309), the Open Research Fund of Key Laboratory of Advanced Theory and Application in Statistics and Data Science (East China Normal University). This work was supported in part by "Pioneer" and "Leading Goose" R\&D Program of Zhejiang (2025C02037) and the Key R\&D Program of Hangzhou (2025SZDA0254).

\bibliography{custom}

\clearpage
\appendix


\input{section/appendices}
\end{document}

%% file: section/introduction.tex
\section{Introduction}
The burgeoning field of artificial intelligence has drawn significant research focus to multi-turn conversational capabilities of large language models (LLMs)~\citep{kwon2024llms,duan2024botchat,deshpande2025multichallenge}. As models grow in scale and complexity, a critical challenge lies in evaluating their ability to maintain coherent and meaningful interactions in dynamic dialogue contexts. To systematically address this, a series of multi-turn dialogue benchmarks has been introduced. Notable examples include MT-Bench~\citep{bai2024mt}, which utilizes powerful LLMs like GPT-4o~\citep{hurst2024gpt} as evaluators for open-domain conversations, and AlpacaEval~\citep{li2023alpacaeval}, which employs automated metrics to rapidly assess model proficiency in multi-turn instruction-following.

However, while these general-purpose benchmarks are useful, they do not fully capture the deeper conversational capabilities of LLMs in vertical domains. As LLM-based agents~\citep{xi2025agentgym,cai2025rtbagent,shangagentsquare} and systems play an increasingly important role in education, a competent LLM must not only maintain coherent dialogues but also exhibit core teaching abilities, such as guiding student reasoning, correcting errors, and providing personalized feedback\textemdash skills that current general benchmarks fail to assess adequately. Consequently, developing a benchmark that systematically evaluates LLMs' multi-turn conversational abilities in domain-specific contexts, particularly education, has become urgent and essential. Figure 1 shows (a) represents single-turn teaching, where the model merely delivers knowledge, whereas (b) depicts multi-turn teacher-student dialogue, in which the teacher actively guides the student through interactive instruction.

\begin{figure*}[tpb]
\centering
  \includegraphics[height=0.52\textwidth]{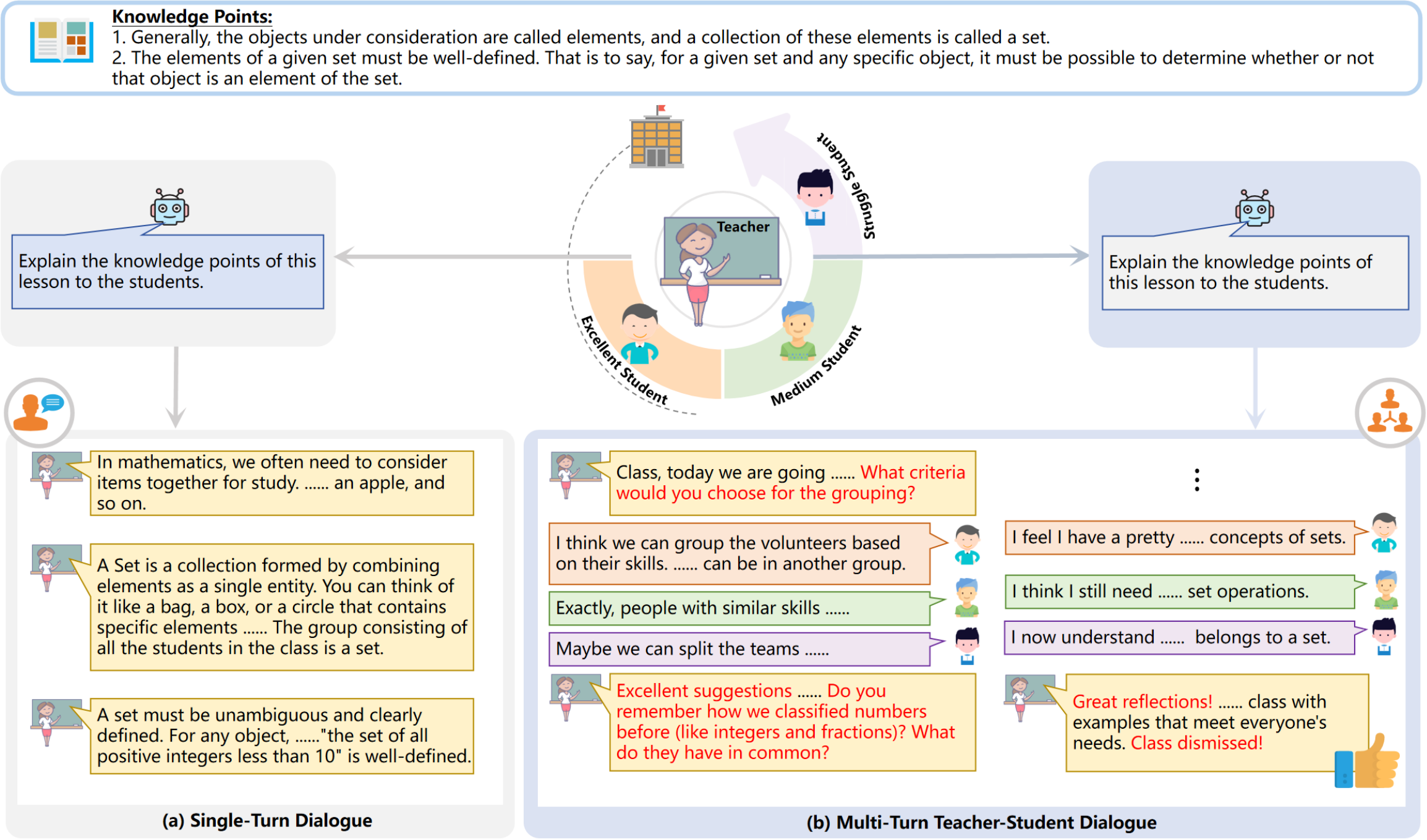}
  \caption{Comparison of ``(a) single-turn ” and ``(b) multi-turn teacher-student” dialogue.}
  \label{fig:fig1}
  \vspace{-2mm}
\end{figure*}

However, constructing a multi-turn teacher-student dialogue benchmark still faces two major challenges:
\textbf{(1) How to accurately determine the optimal timing for teacher questioning?} Inappropriate timing can interrupt students' comprehension, disrupt their reasoning process, and ultimately hinder learning outcomes. For instance, when students are just encountering new knowledge and have not yet grasped it, premature questioning can exacerbate their confusion and significantly reduce learning efficiency. \textbf{(2) How to adjust questioning strategies according to different teaching stages?} Existing approaches often lack stage-specific strategies. Even with proper timing, unsuitable strategies may backfire and negatively impact learning. For example, posing questions beyond a student’s cognitive level at the initial stage of learning can easily lead to frustration and hinder further progress.

To address these challenges, we propose \textbf{EduDial}, a multi-turn teacher-student dialogue dataset covering 345 core knowledge points and comprising 34,250 dialogue sessions generated through teacher-student interactions, designed to enable LLMs to pose appropriate questions at the right time in education. Specifically, to tackle challenge (1), we construct a five-stage progressive teaching process (Introduction, Concept Exploration, Deep Understanding, Knowledge Application, Reflection and Summary) based on Bloom's taxonomy~\cite{forehand2010bloom}, where the completion of each stage indicates that students have digested the current level of knowledge, and posing questions at this moment naturally guides them to the next level, thus each stage represents an appropriate questioning timing. To address challenge (2), we design two questioning strategies for each teaching stage by analyzing MOOC instructional videos and conducting in-depth discussions with experienced mathematics teachers. These strategies, including Scenario-Based Questioning, Zone of Proximal Development Questioning, and Metacognitive Questioning, facilitate the better simulation of authentic classroom interactions. Moreover, we design differentiated teaching strategies for students at varying cognitive levels, providing more targeted guidance. Building on EduDial, we further train \textbf{EduDial-LLM 32B} and propose an 11-dimensional evaluation framework that systematically measures LLMs' teaching capabilities, encompassing both overall teaching quality and content quality. Extensive experiments demonstrate the challenging nature of EduDial and the effectiveness of our approach in enhancing the performance of LLMs in educational tasks.

\begin{figure*}[t]
\centering
  \includegraphics[height=0.65\textwidth]{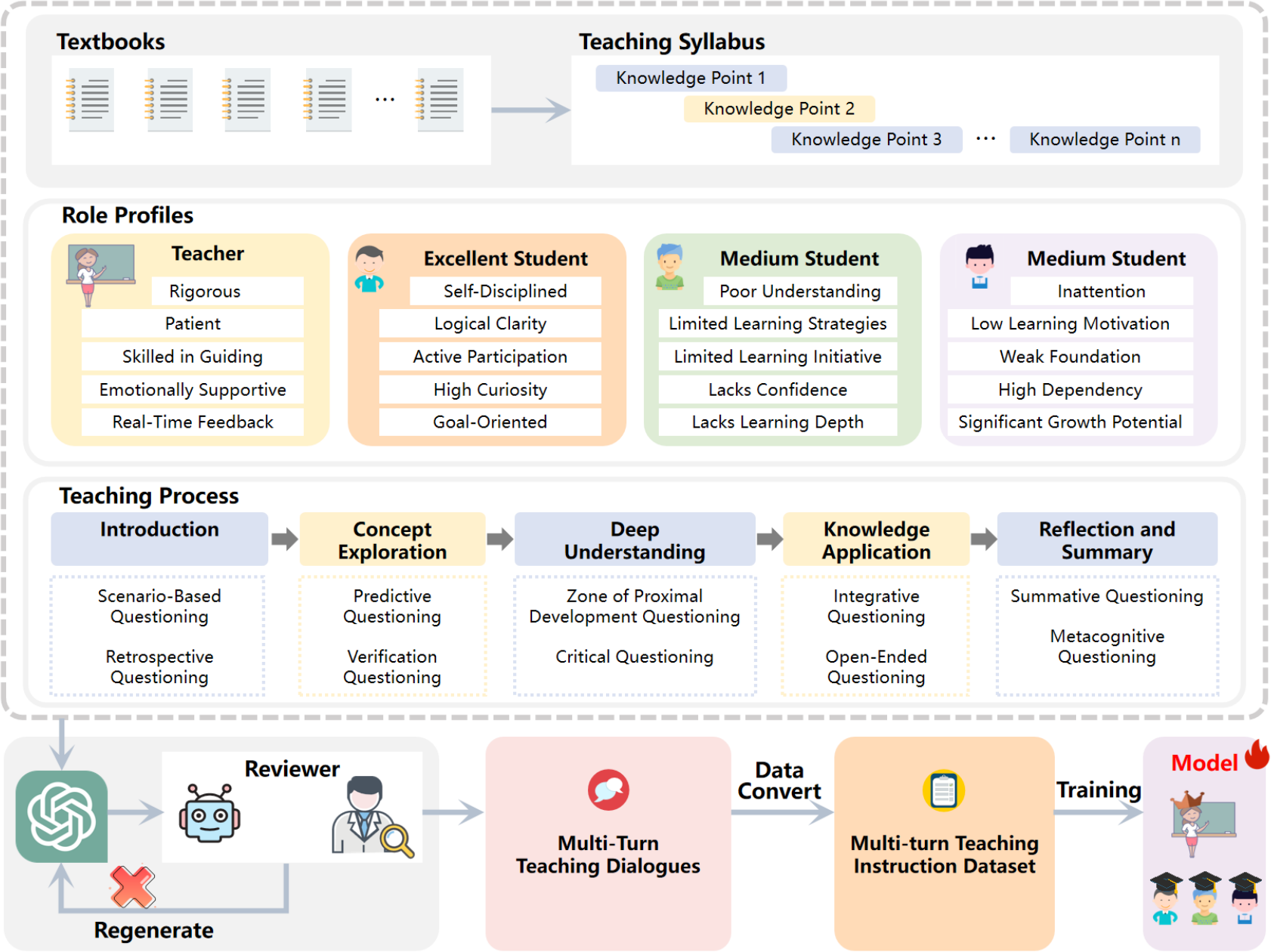} 
  \caption{Based on the teaching syllabus and role profiles, generate high-quality multi-turn teaching instruction data through a five-stage teaching process to train one teacher and three student models.}
  \label{fig:f1}
\end{figure*}

%% file: section/relatedwork.tex
\section{Related Work}
\begin{figure*}[t]
\centering
  \includegraphics[height=0.5\textwidth]{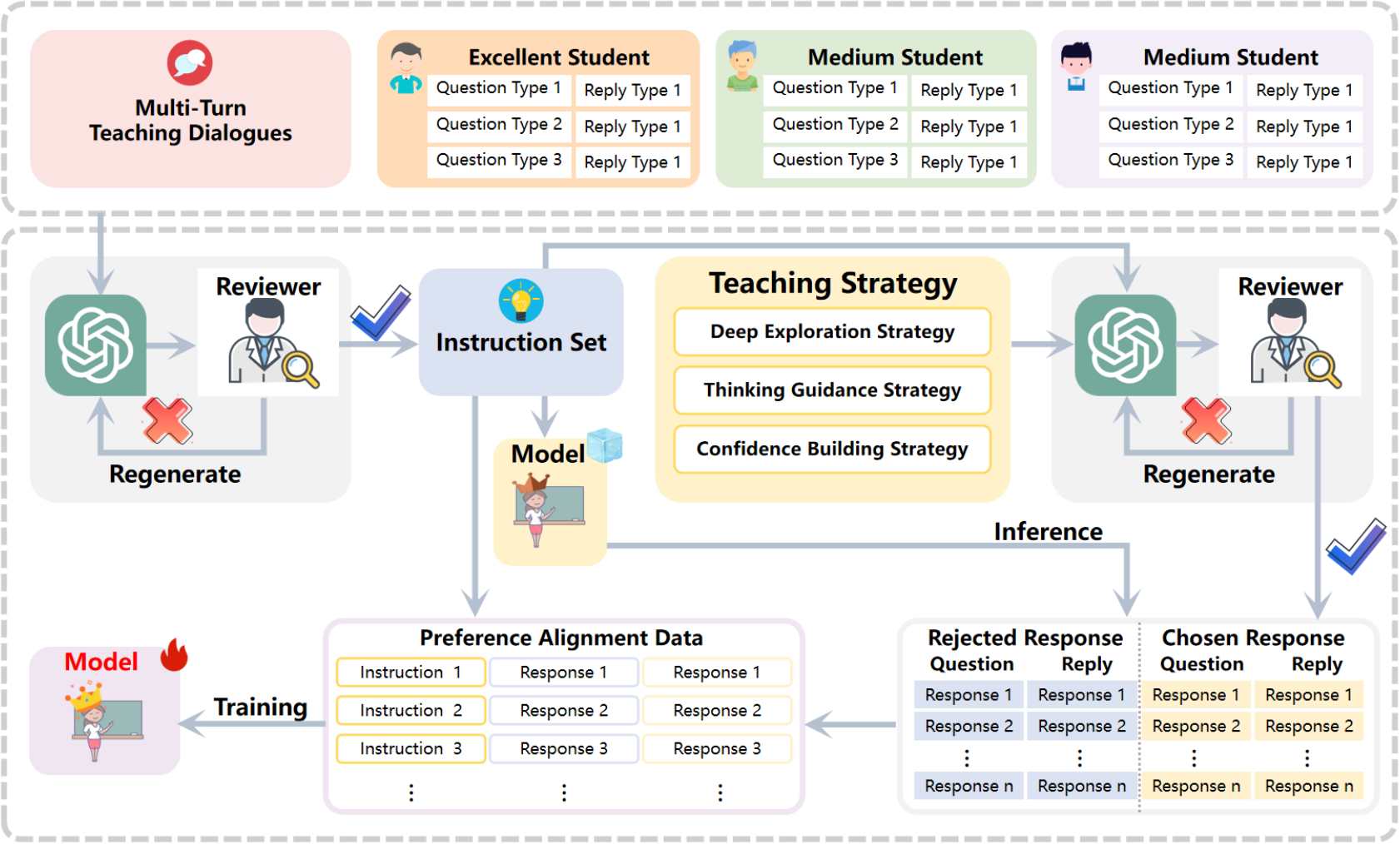} 
  \caption{Develop teaching strategies based on the three student response types, construct preference datasets by generating chosen and rejected responses, and optimize the teacher model using direct preference optimization.}
  \label{fig:f2}
  \vspace{-4mm}
\end{figure*}

\textbf{Dialogue Dataset.}
Early dialogue research focuses on single-turn scenarios, constructing domain-specific datasets for healthcare, finance, and code generation~\citep{zeng2020meddialog, chen2021finqa, dong2024abilities,castlejia2026}. However, single-turn dialogues cannot capture dialogue history understanding and adaptive response strategies, restricting LLM deployment.
Researchers shift toward multi-turn systems. General methods establish benchmarks but lack domain adaptability~\citep{wu2025instruct, bai2024mt}. Domain-specific methods construct datasets in law, medicine, and mathematics through role simulation~\citep{yue2025multi, wang2024notechat, liu2025storm}, but focus on knowledge transfer accuracy rather than guiding student thinking through interaction.
Researchers increasingly focus on multi-turn educational dialogue datasets. Early research relies on costly human annotation~\citep{stasaski2020cima, caines2020teacher, suresh2022talkmoves}. Recent studies employ textbook-based generation~\citep{wang2024book2dial}, pedagogical theory integration~\citep{liu2025vcr}, and Socratic teaching simulation~\citep{liu2024socraticlm}.
However, existing datasets employ rigid questioning and ignore cognitive differences. We propose a five-stage framework defining when and how to ask questions, with strategies for varying cognitive levels.

\noindent\textbf{Leveraging LLMs for Advanced Intelligent Education.} LLMs reshape intelligent education in three areas: teaching content generation, learning assessment and feedback, and interactive teaching~\citep{labadze2023role, stamper2024enhancing, wang2024large,codolzhang2025}.
For teaching content, research focuses on using LLMs to generate teaching plans, exercises, and PPTs~\citep{hu2024teaching, li2024generating, xie2025leveraging,qu2026spot}, providing personalized teaching support.
For learning assessment, LLMs evaluate learners' cognitive levels and provide personalized feedback~\citep{meyer2024using}, creating adaptive educational experiences.
Regarding interactive teaching, most relevant to our research, there are two main research directions.
The first direction uses prompt engineering~\citep{liu2023pre} to directly leverage general large language models for teaching guidance. 
Examples include providing support in programming~\citep{kargupta2024instruct}, classroom education~\citep{zhang2024simulating}, and psychological counseling~\citep{qiu2024interactive}.
The second direction enhances the model's multi-turn interactive teaching ability through specialized training and fine-tuning. 
For example, previous work enhances deep interactive teaching abilities by collecting numerous teaching instructions to train models.

%% file: section/method.tex
\section{The EduDial Dataset}
\subsection{Task Definition}


\noindent\textbf{Teacher-student multi-turn dialogue} introduces role asymmetry and objective divergence to multi-turn interactions. Teachers guide and explain while students express understanding and confusion. This requires dynamic pedagogical adaptation: teachers identify comprehension levels and deliver tailored scaffolding, while students demonstrate varying cognitive states from bewilderment to proficiency. Our work focuses on simulating multi-student classroom scenarios where a teacher interacts with multiple students of varying cognitive levels (excellent, medium, struggling).

\subsection{Preparation}
Unlike general-domain data synthesis, educational dialogue generation relies on pedagogical theories, teaching strategies, and domain knowledge. We establish three components in the preparation.

\textit{1) Teaching Syllabus.}
We surveyed over 100 primary and secondary schools and held extensive discussions with experienced mathematics teachers. We selected 345 core knowledge points from the K-12 curriculum, covering 173 chapters. We extracted each knowledge point from textbooks, including concepts, theorems, and examples, to ensure consistency with curriculum standards.

\textit{2) Teaching Process and Question Strategies.}
We propose a five-stage progressive teaching process aligned with Bloom's taxonomy. The introduction stage (remember) activates prior knowledge through contextual and retrospective questioning. The concept exploration stage (understand) facilitates concept internalization via predictive and verification questioning. The deep understanding stage (analyze) cultivates critical thinking using the zone of proximal development and critical questioning. The knowledge application stage (apply and evaluate) enhances practical competencies through integrative and open-ended questioning. The reflection stage (create) promotes knowledge reconstruction with summary and metacognitive questioning. Question strategies are derived from analyzing MOOC videos and discussions with experienced mathematics teachers, ensuring that each teaching phase has appropriate dual question strategies.

\textit{3) Role Profiles and Teaching Strategies.}
We establish a four-member role system: one teacher and three student profiles (excellent, medium, struggling), characterized by cognitive capacity, learning motivation, classroom engagement, learning strategies, and self-regulation skills. This setup mirrors authentic classrooms where teachers simultaneously address diverse learning needs. Based on these profiles, we design differentiated teaching strategies. Excellent students exhibit inquiry-based and exploratory questioning, with accurate responses showing clear explanations and the ability to connect related concepts. We employ a deep exploration strategy fostering higher-order thinking through progressive questioning. Medium students pose application-oriented and procedural questions, with partially correct responses indicating uncertainty and a tendency to seek hints. We implement a thinking guidance strategy supporting their problem-solving and cognitive clarity. Struggling students ask fundamental questions, with answers that often reveal errors or vagueness. We develop a confidence-building strategy that strengthens foundational cognition and learning confidence through gradual knowledge construction.

\subsection{Dataset Construction}

\textbf{Data Generation}
We detail the construction process of our dataset, EduDial, which is composed of two main components: Multi-turn Teaching Instruction (MTI) and the Preference Dataset based on Teaching Strategies (PDTS).

\textit{MTI Dataset Generation.} 
We design a three-step data generation pipeline for multi-turn teacher-student dialogues in the MTI dataset. Inspired by MT-Bench, we transform the teaching process, role profiles, and teaching syllabus into structured prompt templates. Using LLMs (i.e., o1~\citep{jaech2024openai}), we generate multi-turn teaching dialogues maintaining role consistency and logical coherence aligned with authentic teaching scenarios. Second, we collaborate with experts to establish five core teaching principles: teaching stage integrity, comprehensive questioning strategy, consistent role-playing, accurate knowledge delivery, and natural interaction flow. Based on these principles, we ensure that generated dialogues meet teaching standards through expert-machine dual validation. Finally, we convert validated dialogues into role-specific supervised fine-tuning (SFT) corpora. For the teacher model, training data pairs student responses as input with teacher responses as output. For the student model, training data uses teacher and peer responses as input with student responses as output. The MTI dataset enables training teacher models to master the five-stage teaching process while training student models to maintain role consistency across multi-turn interactions.

\textit{PDTS Dataset Generation.} 
The PDTS dataset generation follows a systematic approach based on differentiated teaching strategies. First, we combine student profile descriptions at each cognitive level with multi-turn teaching dialogues to create structured prompts. These prompts guide LLMs to generate student interactions aligned with specific cognitive levels, with generated questions and replies serving as instructions for teacher model input. Second, we input these instructions into a fine-tuned teacher model to generate rejected responses that are reasonable but deviate from predetermined teaching strategies. Third, we combine instructions with corresponding teaching strategies and input them into LLMs to generate chosen responses that accurately reflect designated teaching methods. Fourth, we construct preference alignment triplets $\langle$instruction, chosen response, rejected response$\rangle$ forming the final PDTS dataset for direct preference optimization (DPO) training. PDTS trains the model to select appropriate teaching strategies for students with different cognitive levels, enabling better implementation of differentiated instruction in multi-student classroom scenarios.

\noindent\textbf{Expert-Machine Dual Verification.}
We employ dual verification, combining expert evaluation and machine assessment to ensure data quality. All generated data undergo automatic verification by GPT-4o and manual review by experts. For the MTI dataset, verification criteria include: logical progression of teaching stages, consistency between questioning strategies and objectives, alignment between student responses and cognitive levels, educational value of feedback, and natural dialogue flow. For the PDTS dataset, verification dimensions encompass teaching strategy adaptability and distinction between chosen and rejected responses. For data failing verification, experts and GPT-4o provide feedback for regeneration until the data meet quality standards.

\subsection{Two-stage Training Strategy}
Our main contribution is constructing two datasets.
To validate their effectiveness, we adopt a two-stage training framework: SFT followed by DPO. The two-stage design addresses learning requirements: SFT learns the five-stage teaching process, while DPO teaches which teaching strategies work best for each student type.
We use these methods to ensure performance gains stem from data quality rather than algorithmic innovations.

\noindent\textbf{SFT Stage.} To ensure LLMs better align with role profiles when simulating different roles, we use instruction fine-tuning to train four LLMs.
We train four LLMs using instruction fine-tuning data for different roles with negative log-likelihood loss:
\begin{equation}
\label{eq1}
\mathcal{L}_{sft} = - \sum_{c=1}^{C} \sum_{k=1}^{K_c} \sum_{i=1}^{N_{c,k}} \log P\left(t_{c,k,i} \mid  t_{c,k,<i},\; C_{c,k}\right).
\end{equation}
\begin{equation}
\label{eq2}
C_{c,k} = \left\{\, (input_{c,j}, output_{c,j}) \,\right\}_{j=1}^{k-1} \cup \{input_{c,k}\},
\end{equation}
where $C$ is batch size, $K_c$ is total rounds for the c-th sample, $N_{c,k}$ is the number of tokens output in the k-th round of the c-th sample, $t_{c,k, i}$ is the i-th token in the k-th round of the c-th sample, $t_{c,k,<i}$ is the token sequence before the i-th token in the k-th round of the c-th sample, $C_{c,k}$ is context before the k-th round of the c-th sample, $input_{c,j}$ is input of the j-th round in the c-th sample, and $output_{c,j}$ is output of the j-th round in the c-th sample.

\noindent\textbf{DPO Stage.} We train the fine-tuned model using direct preference optimization, which maximizes the generation probability of chosen responses while minimizing that of rejected responses. The optimization mechanism ensures that the model tends to generate responses conforming to teaching strategies while suppressing responses deviating from teaching strategies. The loss function is:
\begin{equation}
\label{eq3}
\mathcal{L}_{dpo} = -\log \sigma (s_{chosen} - s_{rejected}).
\end{equation}
\begin{equation}
\label{eq4}
s_{chosen} = \log P(chosen \mid instruction).
\end{equation}
\begin{equation}
\label{eq5}
s_{rejected} = \log P(rejected \mid instruction),
\end{equation}
where $s_{chosen}$ and $s_{rejected}$ are chosen and rejected response scores, and $\sigma$ is the sigmoid function.

\subsection{Evaluation Metrics}
We design two metric categories: overall quality and content quality. The overall quality category comprises nine dimensions (insight, response, feedback, thinking, fluency, interactivity, emotional support, adaptability, and goal), grounded in the Classroom Assessment Scoring System (CLASS)~\cite{pianta2008classroom}, a widely adopted classroom observation framework that assesses teaching quality across three domains: Emotional Support, Instructional Support, and Classroom Organization. Our nine dimensions span all three CLASS domains: \emph{Emotional Support} (emotional support and adaptability), \emph{Instructional Support} (insight, thinking, feedback, response, and fluency), and \emph{Classroom Organization} (interactivity and goal). The content quality category further introduces relevance and coverage to measure the alignment between instructional content and the curriculum. Table~\ref{tab:overall-sub} in the Appendix details all eleven dimensions.
Overall quality and relevance use a five-point Likert scale (1--5); coverage is measured as a percentage (0--100\%).
Evaluation combines human and machine methods: five experts score independently, and their average forms the human result. GPT-4o serves as the machine evaluator, scoring the same data five times independently, with the average as the final machine result. Detailed evaluation protocols, including inter-rater reliability analysis and machine-human correlation validation, are provided in Appendix~\ref{sec:appendix_eval_protocols}.

\subsection{Statistics}
We split the dataset 8:2 by chapter into training (137 chapters, 275 knowledge points) and test sets (36 chapters, 70 knowledge points).
We construct two datasets: (1) MTI Dataset: 100 dialogues per chapter (13,700 total), averaging 11 rounds, and teachers pose an average of 14.2 questions per dialogue. (2) PDTS Dataset: 50 instructions per student role, each yielding one preference pair (⟨instruction, chosen, rejected⟩), totaling 20,550 pairs under 1000 tokens each, enabling efficient DPO training.
For evaluation, we create 36 teaching scenarios per model to assess performance in multi-student classroom settings (reported in Table~\ref{tab:combined_evaluation}). Each model acts as the teacher, simultaneously engaging with three students of varying cognitive levels. Sessions end when the teacher completes instruction or after 15 rounds. To further validate model effectiveness across different scenarios, we conduct an additional one-on-one evaluation (reported in Table~\ref{tab:ablation_results}) where the teacher model interacts with a single student, also limited to 15 rounds. Both evaluation settings employ identical assessment dimensions and methodologies. During evaluation, teacher models instruct knowledge points not encountered during training, demonstrating generalization capability.

%% file: section/experiment.tex
\section{Experiments}

\subsection{Experimental Setup}

\textbf{Baselines.}
We select 18 large language models as baselines, including nine closed-source and eight open-source models.
For closed-source models, we evaluate Gemini-2.5-Pro-Exp~\citep{gemini25}, Gemini-2.0-Flash-Exp~\citep{gemini20flash}, Claude-3-7-Sonnet~\citep{claude37sonnet}, Claude-3-7-Sonnet-Thinking~\citep{claude37sonnet}, Claude-3-5-Sonnet~\citep{claude35sonnet}, o3-mini-High~\citep{o3mini}, o1~\citep{jaech2024openai}, GPT-4o~\citep{hurst2024gpt} and GPT-3.5~\citep{ouyang2022training}.
For open-source models, we evaluate DeepSeek-R1~\citep{guo2025deepseek}, DeepSeek-V3~\citep{liu2024deepseek}, Llama-3.3-70B-Instruct~\citep{llama33}, Yi-1.5-34B-Chat~\citep{young2024yi}, Qwen-2.5-72B-Instruct~\citep{yang2024qwen2}, Qwen-2.5-72B-Math~\citep{yang2024qwen2m}, QwQ-32B-Preview~\citep{qwq32b}, SocraticLM~\citep{liu2024socraticlm}, and our EduDial-LLM, which is fine-tuned from QwQ-32B-Preview.

\begin{table*}[t]
\centering
\renewcommand{\arraystretch}{1.2} 
\large 
\setlength{\tabcolsep}{3pt}
\resizebox{\textwidth}{!}{
\begin{tabular}{l|cc cc cc cc cc cc cc cc cc cc| cc cc}
\toprule
\multirow{3}{*}{\textbf{Model}} & \multicolumn{20}{c|}{\textbf{Overall Quality}} & \multicolumn{4}{c}{\textbf{Content Quality}} \\
\cmidrule(lr){2-21} \cmidrule(lr){22-25}
 & \multicolumn{2}{c|}{\textbf{INS}} & \multicolumn{2}{c|}{\textbf{RES}} & \multicolumn{2}{c|}{\textbf{FB}} & \multicolumn{2}{c|}{\textbf{THK}} & \multicolumn{2}{c|}{\textbf{INT}} & \multicolumn{2}{c|}{\textbf{EMO}} & \multicolumn{2}{c|}{\textbf{ADP}} & \multicolumn{2}{c|}{\textbf{FLU}} & \multicolumn{2}{c|}{\textbf{GOL}} & \multicolumn{2}{c|}{\textbf{AVG}} & \multicolumn{2}{c|}{\textbf{REL}} & \multicolumn{2}{c}{\textbf{COV(\%)}} \\
\cmidrule(lr){2-3} \cmidrule(lr){4-5} \cmidrule(lr){6-7} \cmidrule(lr){8-9} \cmidrule(lr){10-11} \cmidrule(lr){12-13} \cmidrule(lr){14-15} \cmidrule(lr){16-17} \cmidrule(lr){18-19} \cmidrule(lr){20-21} \cmidrule(lr){22-23} \cmidrule(lr){24-25}
 & \textbf{M} & \textbf{H} & \textbf{M} & \textbf{H} & \textbf{M} & \textbf{H} & \textbf{M} & \textbf{H} & \textbf{M} & \textbf{H} & \textbf{M} & \textbf{H} & \textbf{M} & \textbf{H} & \textbf{M} & \textbf{H} & \textbf{M} & \textbf{H} & \textbf{M} & \textbf{H} & \textbf{M} & \textbf{H} & \textbf{M} & \textbf{H} \\
\midrule
\multicolumn{25}{c}{\cellcolor{gray!15}\textbf{Closed Source Models}} \\
\midrule
Gemini-2.5-Pro-Exp~\citep{gemini25} & 4.06 & 4.05 & 4.23 & 4.41 & 3.84 & 3.95 & 3.48 & 3.71 & 4.56 & 4.58 & \textbf{4.67} & \textbf{4.64} & 4.12 & 3.97 & \textbf{4.67} & \textbf{4.62} & 4.09 & 4.08 & 4.19 & 4.22 & \textbf{4.95} & \textbf{4.81} & 94.17 & 91.98 \\
\rowcolor{gray!5}
Gemini-2.0-Flash-Exp~\citep{gemini20flash} & 3.89 & 3.78 & 4.34 & 4.37 & 3.89 & 4.00 & 4.29 & 4.41 & 3.57 & 3.51 & 3.69 & 4.01 & 3.89 & 3.79 & 4.46 & 4.64 & 4.31 & 4.29 & 4.04 & 4.09 & 4.94 & 4.78 & \textbf{96.16} & \textbf{94.84} \\
Claude-3-7-Sonnet~\citep{claude37sonnet} & 3.90 & 3.93 & 4.33 & 4.29 & 3.67 & 3.84 & 3.39 & 3.51 & 4.36 & 4.37 & 4.39 & 4.32 & 3.91 & 3.93 & 4.50 & 4.59 & 4.01 & 3.98 & 4.05 & 4.08 & 4.89 & 4.76 & 90.08 & 88.47 \\
\rowcolor{gray!5}
Claude-3-7-Sonnet-Thinking~\citep{claude37sonnet} & 4.05 & 3.91 & \textbf{4.63} & \textbf{4.50} & 3.85 & 3.71 & 4.19 & 4.08 & 3.42 & 3.11 & 3.28 & 3.34 & 3.63 & 3.53 & 4.63 & 4.61 & 4.29 & 4.29 & 4.00 & 3.90 & 4.60 & 4.49 & 90.26 & 89.31 \\
Claude-3-5-Sonnet~\citep{claude35sonnet} & 3.82 & 3.71 & 4.29 & 4.00 & 3.68 & 3.40 & 3.44 & 3.38 & 3.62 & 3.21 & 3.97 & 3.86 & 3.79 & 3.49 & 4.47 & 4.14 & 3.94 & 4.05 & 3.89 & 3.69 & 4.80 & 4.68 & 93.70 & 91.82 \\
\rowcolor{gray!5}
o3-mini-High~\citep{o3mini} & 3.66 & 3.45 & 4.24 & 4.10 & 3.41 & 3.18 & 3.23 & 3.30 & 2.86 & 2.60 & 3.40 & 3.23 & 2.91 & 2.78 & 4.42 & 4.30 & 3.76 & 3.77 & 3.54 & 3.41 & 4.20 & 4.12 & 80.11 & 79.24 \\
o1~\citep{jaech2024openai} & 3.50 & 3.36 & 4.30 & 4.30 & 3.20 & 3.17 & 3.42 & 3.51 & 2.66 & 2.62 & 3.04 & 2.86 & 2.82 & 2.71 & 4.40 & 4.34 & 3.91 & 3.91 & 3.47 & 3.42 & 4.54 & 4.46 & 90.17 & 92.91 \\
\rowcolor{gray!5}
GPT-4o~\citep{hurst2024gpt} & 3.75 & 3.63 & 4.14 & 4.08 & 3.61 & 4.04 & 3.92 & 3.92 & 3.28 & 3.00 & 3.47 & 2.80 & 3.47 & 2.86 & 4.31 & 4.27 & 3.81 & 4.20 & 3.75 & 3.64 & 4.83 & 4.73 & 88.52 & 84.89 \\
GPT-3.5~\citep{ouyang2022training} & 3.67 & 3.50 & 4.03 & 3.85 & 3.22 & 3.10 & 3.92 & 3.84 & 2.92 & 2.58 & 3.25 & 2.56 & 3.28 & 2.94 & 4.31 & 3.93 & 3.72 & 3.86 & 3.59 & 3.35 & 4.76 & 4.61 & 91.30 & 89.83 \\
\midrule
\multicolumn{25}{c}{\cellcolor{gray!15}\textbf{Open Source Models}} \\
\midrule
\rowcolor{gray!5}
Deepseek-R1~\citep{guo2025deepseek} & 3.77 & 3.54 & 4.50 & 4.42 & 3.61 & 3.44 & 3.79 & 4.03 & 2.73 & 2.62 & 2.76 & 2.58 & 3.40 & 3.27 & 4.18 & 4.23 & 4.16 & 4.17 & 3.66 & 3.59 & 4.57 & 4.48 & 90.18 & 89.93 \\
Deepseek-V3~\citep{liu2024deepseek} & 3.39 & 3.25 & 4.09 & 4.26 & 3.21 & 3.09 & 3.08 & 3.25 & 2.50 & 2.31 & 2.72 & 2.58 & 2.77 & 2.74 & 4.07 & 4.35 & 3.94 & 3.91 & 3.31 & 3.30 & 4.62 & 4.51 & 95.04 & 94.32 \\
\rowcolor{gray!5}
Llama-3.3-70B-Instruct~\citep{llama33} & 2.56 & 2.04 & 3.08 & 1.97 & 2.53 & 2.50 & 2.36 & 2.25 & 2.81 & 2.15 & 2.72 & 2.23 & 2.47 & 2.06 & 4.25 & 3.47 & 3.56 & 3.36 & 2.93 & 2.45 & 4.33 & 4.12 & 88.19 & 87.69 \\
Yi-1.5-34B-Chat~\citep{young2024yi} & 2.28 & 2.10 & 3.11 & 3.17 & 2.81 & 2.86 & 2.28 & 2.28 & 2.75 & 2.92 & 2.89 & 3.08 & 2.44 & 2.78 & 3.64 & 3.22 & 3.58 & 3.81 & 2.86 & 2.91 & 4.28 & 4.11 & 86.92 & 85.53 \\
\rowcolor{gray!5}
Qwen-2.5-72B-Instruct~\citep{yang2024qwen2} & 3.18 & 2.89 & 3.45 & 2.74 & 2.91 & 2.43 & 2.70 & 2.89 & 2.59 & 2.78 & 2.97 & 2.67 & 2.24 & 2.41 & 3.30 & 3.62 & 2.97 & 3.39 & 2.92 & 2.87 & 4.48 & 4.27 & 89.42 & 87.75 \\
Qwen-2.5-72B-Math~\citep{yang2024qwen2m} & 2.02 & 3.06 & 2.32 & 3.60 & 1.96 & 2.63 & 1.93 & 2.58 & 1.91 & 2.57 & 1.89 & 2.44 & 1.84 & 2.01 & 2.84 & 3.26 & 2.70 & 3.19 & 2.16 & 2.82 & 4.35 & 4.16 & 83.89 & 82.43 \\
\rowcolor{gray!5}
QwQ-32B-Preview~\citep{qwq32b} & 2.89 & 2.22 & 3.24 & 2.42 & 2.65 & 2.28 & 2.43 & 2.33 & 2.69 & 2.25 & 2.67 & 2.31 & 2.16 & 2.22 & 3.15 & 3.44 & 2.89 & 3.31 & 2.75 & 2.53 & 4.43 & 4.25 & 88.08 & 86.56 \\
SocraticLM~\citep{liu2024socraticlm} & 2.69 & 2.97 & 3.69 & 3.96 & 3.31 & 3.48 & 3.08 & 3.07 & 3.56 & 3.46 & 3.56 & 3.43 & 2.83 & 2.82 & 3.58 & 3.94 & 4.03 & 3.81 & 3.37 & 3.44 & 4.39 & 4.31 & 87.94 & 86.31 \\
\midrule
\rowcolor{blue!10} EduDial-LLM (Ours) & \textbf{4.58} & \textbf{4.43} & 4.12 & 4.26 & \textbf{4.62} & \textbf{4.25} & \textbf{4.55} & \textbf{4.41} & \textbf{4.64} & \textbf{4.68} & 4.10 & 4.09 & \textbf{4.60} & \textbf{4.25} & 4.43 & 4.49 & \textbf{4.56} & \textbf{4.42} & \textbf{4.47} & \textbf{4.36} & 4.83 & 4.77 & 91.83 & 88.29 \\
\bottomrule
\end{tabular}}
\caption{Comprehensive Machine (M) and Human (H) Evaluation of 18 Models across Overall and Content Quality.}
\label{tab:combined_evaluation}
\end{table*}
\noindent\textbf{Experiments Settings.} 
We conduct experiments on two NVIDIA A100 80GB GPUs with CUDA 12.4, PyTorch 2.4.0, and Python 3.10.
For SFT, we apply QLoRA~\citep{dettmers2023qlora} with 4-bit quantization (rank r=64, $\alpha$=16), AdamW optimizer, and BF16 mixed precision, evaluating every 500 steps. 
Each epoch in this stage takes approximately 3 hours to complete, with a total computation time of 45 hours across all four models. 
For DPO, we use a learning rate of 5e-7, batch size 8, gradient accumulation over 16 steps, and train for two epochs with the same QLoRA configuration. 
We set $\beta$=0.1 and the maximum sequence length to 1024 tokens.
Each epoch requires approximately 4 hours, resulting in a total computation time of 8 hours for this stage. 
Additional hyperparameters are detailed in Table~\ref{tab:training-config} in the appendix. All values are determined via grid search on the test set.

\subsection{Performance Comparison}

Our model achieves the highest average scores in both machine (4.47) and human (4.36) evaluations, with balanced performance across all nine quality dimensions.
Baselines, despite strong general capabilities, exhibit clear strengths and weaknesses. Closed-source models such as Gemini-2.5-Pro-Exp perform well on content quality (e.g., REL: 4.95) and dialogue quality (e.g., emotional support: 4.67, fluency: 4.67), but fall short on core pedagogical dimensions. We identify two failure patterns:
(1) \textbf{Lack of guided teaching.} Models such as DeepSeek-V3 (THK: 3.08, INT: 2.50) tend to provide answers directly rather than guiding step-by-step reasoning, undermining the objective of fostering independent thinking.
(2) \textbf{Role consistency collapse.} Several baselines maintain the teacher role in early turns but degrade into general QA systems as the dialogue progresses, ceasing to pose follow-up questions or advance teaching stages.
Open-source models generally score below 4.0 across multiple dimensions, reflecting limited pedagogical capability without task-specific training.

EduDial-LLM overcomes these limitations through our two-stage training paradigm with carefully constructed datasets.
The MTI dataset used in the SFT stage endows the model with structured pedagogical objectives and reliable content processing capabilities, yielding strong performance in goal-oriented teaching (4.56), relevance (4.83), and coverage (91.83\%).
Through the five-stage teaching framework (detailed in Appendix Table~\ref{tab:questioning_strategies}), the model masters sophisticated questioning techniques that promote teacher-student interaction and stimulate critical thinking, achieving high scores in thinking (4.55) and insight (4.58).
The PDTS dataset used in the DPO stage further enhances the model's ability to perceive varied student states and adapt teaching strategies in real time. By learning from three distinct teaching strategies (detailed in Appendix Table~\ref{tab:teaching_strategies}), the model provides personalized guidance based on student responses, leading to superior performance in interactivity (4.64), adaptability (4.60), and feedback (4.62).

\subsection{Ablation Study}

\begin{figure}[htpb]
    \centering
    \includegraphics[width=7.5cm]{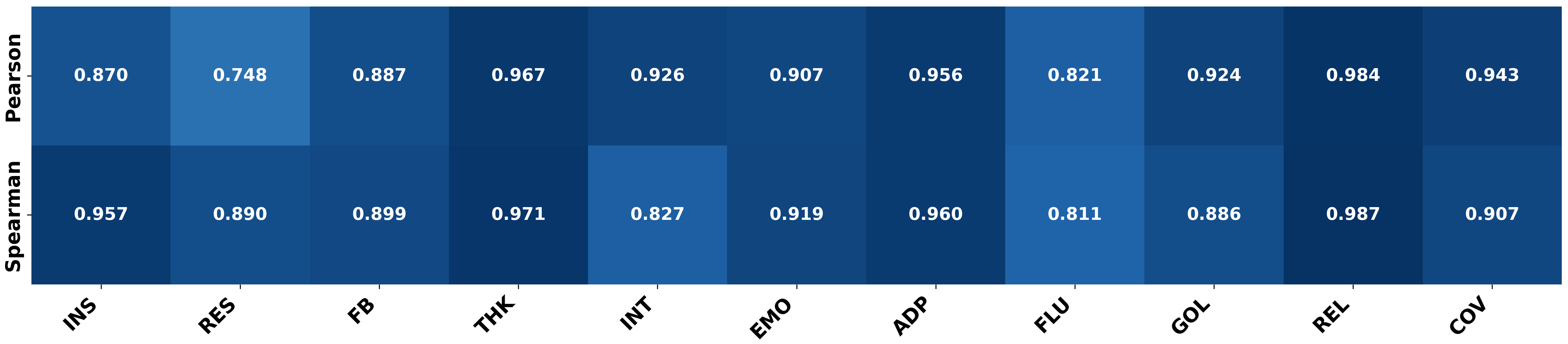}
   \caption{Machine-Human Rating Consistency: Pearson and Spearman Correlations for 18 Models.}
    \label{fig:example}
    \vspace{-2mm}
\end{figure}

\noindent\textbf{Human-Machine Rating Consistency Analysis.} 
As shown in Fig.~\ref{fig:example}, we evaluated consistency between human and machine ratings across 11 dimensions using Pearson and Spearman correlation coefficients. 
The results demonstrate robust inter-rater consistency with average Pearson and Spearman coefficients of 0.90 and 0.91, respectively.
Analysis of all dimensions shows content quality aspects 
(Relevance and Coverage) consistently achieve high correlation coefficients exceeding 0.90.
In contrast, the nine overall quality dimensions exhibit variable performance. 
While thinking and adaptability demonstrate high consistency, interactivity, feedback, and insight show comparatively lower agreement.
While machine evaluation systems align strongly with human expert assessment for dimensions with explicit criteria and objective metrics, they show decreased agreement on dimensions requiring complex contextual understanding and teaching interaction quality assessment.
Despite these variations, the overall high consistency indicates machine rating reliability. 
Therefore, we employ machine ratings as the primary assessment method in subsequent experiments.

\begin{table}[t]
\centering
\resizebox{\columnwidth}{!}{%
\renewcommand{\arraystretch}{0.9}
\begin{tabular}{l|cccccccccc|cc}
\toprule
\multirow{2}{*}{\textbf{Model}} & \multicolumn{10}{c|}{\textbf{Overall Quality}} & \multicolumn{2}{c}{\textbf{Content Quality}} \\
\cmidrule(lr){2-11} \cmidrule(lr){12-13}
 & \textbf{INS} & \textbf{RES} & \textbf{FB} & \textbf{THK} & \textbf{INT} & \textbf{EMO} & \textbf{ADP} & \textbf{FLU} & \textbf{GOL} & \textbf{AVG} & \textbf{REL} & \textbf{COV(\%)} \\
\midrule
\multicolumn{13}{c}{\cellcolor{gray!15}\textbf{Complete Model (QwQ-32B-Preview + SFT + DPO)}} \\
\midrule
\rowcolor{blue!10} EduDial-LLM (E) & \textbf{4.28} & \textbf{4.49} & \textbf{4.61} & \textbf{4.27} & 4.31 & 4.12 & \textbf{4.78} & 4.70 & \textbf{4.86} & \textbf{4.49} & \textbf{4.78} & \textbf{91.83} \\
\rowcolor{blue!10} EduDial-LLM (M) & 4.23 & 4.41 & 4.52 & 4.24 & 4.25 & \textbf{4.19} & 4.64 & \textbf{4.89} & 4.75 & 4.46 & 4.68 & 90.87 \\
\rowcolor{blue!10} EduDial-LLM (S) & 4.27 & 4.46 & 4.59 & 4.25 & \textbf{4.42} & 4.18 & 4.58 & 4.77 & 4.83 & 4.48 & 4.74 & 91.42 \\
\midrule
\multicolumn{13}{c}{\cellcolor{gray!15}\textbf{SFT Only (QwQ-32B-Preview + SFT)}} \\
\midrule
EduDial-LLM w/o DPO (E) & 4.16 & 4.40 & 4.31 & 3.92 & 4.18 & 3.94 & 4.17 & 4.83 & 4.53 & 4.27 & 4.63 & 89.03 \\
\rowcolor{gray!5} EduDial-LLM w/o DPO (M) & 4.14 & 4.43 & 4.38 & 3.81 & 4.19 & 3.98 & 4.14 & 4.79 & 4.47 & 4.26 & 4.64 & 88.31 \\
EduDial-LLM w/o DPO (S) & 4.08 & 4.45 & 4.32 & 3.89 & 4.14 & 3.91 & 4.13 & 4.77 & 4.45 & 4.24 & 4.66 & 89.14 \\
\midrule
\multicolumn{13}{c}{\cellcolor{gray!15}\textbf{DPO Only (QwQ-32B-Preview + DPO)}} \\
\midrule
\rowcolor{gray!5} EduDial-LLM w/o SFT (E) & 3.21 & 3.83 & 3.95 & 3.99 & 3.96 & 3.65 & 3.72 & 4.02 & 3.98 & 3.81 & 4.45 & 87.25 \\
EduDial-LLM w/o SFT (M) & 3.15 & 3.72 & 3.56 & 3.93 & 3.82 & 3.51 & 3.64 & 3.85 & 4.12 & 3.70 & 4.38 & 86.34 \\
\rowcolor{gray!5} EduDial-LLM w/o SFT (S) & 3.29 & 3.89 & 3.71 & 3.87 & 3.91 & 3.69 & 3.75 & 3.93 & 3.78 & 3.76 & 4.43 & 89.05 \\
\midrule
\multicolumn{13}{c}{\cellcolor{gray!15}\textbf{Base Model (QwQ-32B-Preview)}} \\
\midrule
EduDial-LLM w/o SFT w/o DPO (E) & 2.67 & 3.11 & 3.33 & 2.56 & 3.33 & 3.31 & 2.33 & 3.94 & 3.33 & 3.10 & 4.39 & 89.83 \\
\rowcolor{gray!5} EduDial-LLM w/o SFT w/o DPO (M) & 2.28 & 3.06 & 2.44 & 2.17 & 3.00 & 2.86 & 2.86 & 3.44 & 3.78 & 2.88 & 4.29 & 82.47 \\
EduDial-LLM w/o SFT w/o DPO (S) & 2.83 & 3.17 & 2.97 & 2.14 & 3.14 & 3.22 & 2.61 & 3.75 & 3.06 & 2.99 & 4.37 & 88.58 \\
\bottomrule
\end{tabular}
}
\caption{EduDial-LLM One-on-One Teaching Performance Across Configurations and Proficiency Levels. Students: Excellent (E), Medium (M), Struggling (S).}
\label{tab:ablation_results}
\end{table}

\noindent\textbf{Impact of Training Strategies on Teaching Capabilities.} 
Table~\ref{tab:ablation_results} shows evaluation results across 11 dimensions for the complete model and three ablation models when teaching students of different performance levels.
The experimental results demonstrate:
In overall quality, the complete model achieves optimal performance across all student types, significantly outperforming other ablation models.
Among single-stage training approaches, SFT significantly enhances base model performance, effectively teaching the model questioning techniques. 
Similarly, DPO notably strengthens thinking and adaptability dimensions while yielding modest interactivity gains, indicating its efficacy for complex teaching capabilities.
Moreover, two-stage training optimizes performance beyond SFT alone, primarily bolstering thinking and adaptability dimensions.
Analysis across different performance levels shows that the complete model possesses excellent adaptive capabilities, providing personalized teaching according to different learning needs.
Content quality assessment exhibits minimal fluctuation across all models.
However, the complete model still achieves notable improvements.
Experimental results prove that two-stage training ensures accuracy and comprehensiveness in knowledge transfer during teaching.

\noindent\textbf{Additional Experimental Analyses.}
We conduct additional analyses in the Appendix to validate our approach.
First, we conduct zero-shot cross-dataset evaluation on the SocraTeach benchmark to verify pedagogical transferability (Table~\ref{tab:cross_dataset}).
Second, we evaluate training corpus quality through  (Table~\ref{tab:corpus_quality}): (1) SFT data assessment with different datasets (SocraticLM vs. MTI); (2) DPO validation through reversed preference pairs; and (3) cross-dataset robustness.
Third, we examine backbone selection by training multiple architectures (Table~\ref{tab:backbone_selection}).
Fourth, we assess out-of-distribution generalization with six distinct student types (Table~\ref{tab:ood_generalization}).
Fifth, we verify simulated student quality through expert evaluation across five cognitive dimensions (Table~\ref{tab:student_simulation_quality}).
Finally, we evaluate mathematical reasoning capabilities on Math500 and AIME2024 benchmarks (Table~\ref{tab:math_benchmarks}).
These analyses validate our dataset construction, training methodology, and evaluation framework.

%% file: section/conclusion.tex
\section{Conclusion}
In this work, we introduce EduDial-LLM, a novel large language model tailored for multi-turn teacher-student dialogue teaching. Departing from the traditional single-turn paradigm, EduDial-LLM actively engages students through a structured questioning process, emulating effective pedagogical practices found in real-world classrooms. Central to our approach is the EduDial dataset, which supports two-stage training through instructional fine-tuning and preference optimization, enabling the model to adapt its teaching strategies to diverse learner profiles. To rigorously evaluate teaching quality, we also propose a comprehensive 11-dimensional evaluation framework. Experimental results on 17 mainstream LLMs validate the effectiveness of our approach: while most models struggle with student-driven interaction, EduDial-LLM consistently achieves superior performance across all evaluation dimensions. This work underscores the value of guided, interactive teaching in intelligent education and offers a new direction for future LLM-based tutoring systems.

%% file: section/appendices.tex
\newpage
\section{Appendix}
\subsection{Experimental Configuration and Evaluation Metrics}
\label{sec:appendix_eval_protocols}
\noindent\textbf{Evaluation Protocol.}
We employ a dual evaluation framework combining machine-based assessment (GPT-4o) and human expert evaluation.
Five experienced mathematics teachers independently evaluate teaching quality across 11 dimensions defined in Table~\ref{tab:overall-sub}.
To ensure evaluation reliability, we measure inter-rater consistency using Cohen's Kappa coefficient, achieving an average Kappa of 0.76, indicating substantial agreement among experts.
Machine-human correlation analysis (detailed in §4.3) demonstrates strong consistency with Pearson and Spearman coefficients of 0.90 and 0.91, respectively, validating the use of machine evaluation as the primary assessment method for efficiency.

\begin{table}[!ht]
\centering
\resizebox{\columnwidth}{!}{%
\renewcommand{\arraystretch}{1.1}
\begin{tabular}{p{3.5cm}p{0.8cm}p{14cm}}
\toprule
\rowcolor{gray!15} \textbf{Dimension} & \textbf{Abbr.} & \textbf{Definition} \\
\midrule
\multicolumn{3}{l}{\textbf{\textit{Overall Quality}}} \\
\addlinespace[2pt]
Insight & INS & Assess whether the teacher accurately identifies student learning needs, knowledge levels, and question intent. \\ 
\rowcolor{gray!5}
Response & RES & Assess whether the teacher effectively addresses student questions and provides feasible, constructive guidance. \\ 
Feedback & FB & Assess whether the teacher provides timely and effective feedback that facilitates improvement. \\ 
\rowcolor{gray!5}
Thinking & THK & Assess whether the teacher fosters thinking skills, including analysis and open-mindedness. \\ 
Fluency & FLU & Assess whether the teacher communicates clearly and is easy to understand. \\ 
\rowcolor{gray!5}
Interactivity & INT & Assess the frequency and quality of interactions between teacher and students in teaching dialogues. \\ 
Emotional Support & EMO & Assess whether the teacher provides emotional support during teaching and creates a positive learning environment. \\ 
\rowcolor{gray!5}
Adaptability & ADP & Assess whether the teacher adjusts teaching methods to accommodate different contexts, ensuring personalized guidance. \\ 
Goal & GOL & Assess whether teacher guidance helps achieve teaching goals, such as knowledge acquisition. \\
\midrule
\multicolumn{3}{l}{\textbf{\textit{Content Quality}}} \\
\addlinespace[2pt]
\rowcolor{gray!5}
Relevance & REL & Assess whether the teacher-student dialogue is relevant to the teaching content of this chapter. \\ 
Coverage & COV & Assess the proportion of required teaching knowledge points encompassed within the teacher-student dialogue. \\
\bottomrule
\end{tabular}
}
\caption{Evaluation Metrics.}
\label{tab:overall-sub}
\end{table}

\noindent\textbf{Training Configuration.}
We train four models in our framework: one teacher model and three student models representing different proficiency levels. The teacher model is based on QwQ-32B-Preview, selected for its strong reasoning capabilities, while student models use the Qwen2.5 series with varying scales (32B, 14B, and 7B) to simulate different cognitive abilities. Table~\ref{tab:training-config} presents the hyperparameter settings for each model during the supervised fine-tuning (SFT) stage. We employ differentiated learning rates and training epochs to account for the varying complexities of their roles, with the teacher model requiring fewer epochs due to its larger base capacity, while struggling student models benefit from extended training.

\begin{table}[!ht]
\centering

\resizebox{\columnwidth}{!}{%
\renewcommand{\arraystretch}{1.0}
\begin{tabular}{l|c|c|c|c}
\toprule
\rowcolor{gray!15}
\textbf{Parameter} & \textbf{Teacher} & \textbf{Excellent Student} & \textbf{Medium Student} & \textbf{Struggling Student} \\
\midrule
Base Model & QwQ-32B-Preview & Qwen2.5-32B-Instruct & Qwen2.5-14B-Instruct & Qwen2.5-7B-Instruct \\
\rowcolor{gray!5}
Learning Rate & 3e-4 & 5e-4 & 2e-4 & 5e-5 \\
Batch Size & 16 & 16 & 32 & 32 \\
\rowcolor{gray!5}
Epochs & 2 & 3 & 5 & 5 \\
\bottomrule
\end{tabular}
}
\caption{Parameter Settings in SFT for Four Models.}
\label{tab:training-config}
\end{table}

\subsection{Ablation Studies}

\subsubsection{Cross-Dataset Generalization}
\label{app:cross_dataset}
To evaluate whether the pedagogical capabilities of EduDial-LLM generalize beyond its training distribution, we conduct a zero-shot evaluation on the SocraTeach benchmark~\cite{liu2025socraticlm}, built upon math word problems from GSM8K/MAWPS with an independent five-dimensional evaluation framework. EduDial-LLM was not exposed to any SocraTeach data during training. We strictly follow the original evaluation protocol: 100 samples each for IARA, CARA, SER, and SRR, while Overall Quality uses 1,000 dialogues for pairwise blind comparison against GPT-4. We recruit 3 annotators with educational backgrounds for human evaluation, and measure inter-annotator agreement via Fleiss' Kappa, obtaining a score of 0.63, indicating substantial agreement.

\begin{table}[!ht]
\centering
\resizebox{\columnwidth}{!}{%
\renewcommand{\arraystretch}{1.1}
\begin{tabular}{l|ccccc}
\toprule
\rowcolor{gray!15}
\textbf{Model} & \textbf{Overall} & \textbf{IARA} & \textbf{CARA} & \textbf{SER} & \textbf{SRR} \\
\midrule
\multicolumn{6}{l}{\cellcolor{gray!15}\textit{Reported Results}} \\
\midrule
GPT-4 & 0.50 & 0.76 & 0.91 & 0.65 & 0.55 \\
\rowcolor{gray!5}
SocraticLM$^\dagger$ & 0.62 & 0.83 & 0.98 & 0.74 & 0.78 \\
\midrule
\multicolumn{6}{l}{\cellcolor{gray!15}\textit{Our Evaluation}} \\
\midrule
QwQ-32B-Preview (backbone) & 0.31 & 0.59 & 0.77 & 0.48 & 0.26 \\
\rowcolor{gray!5}
EduDial-LLM (Ours) & \textbf{0.57} & \textbf{0.78} & \textbf{0.93} & \textbf{0.70} & \textbf{0.63} \\
\bottomrule
\end{tabular}
}
\caption{Zero-shot cross-dataset evaluation on the SocraTeach benchmark. $^\dagger$SocraticLM is trained on SocraTeach data, while EduDial-LLM is evaluated in a fully zero-shot setting.}
\label{tab:cross_dataset}
\end{table}

As shown in Table~\ref{tab:cross_dataset}, the results reveal three findings.
First, EduDial-LLM surpasses GPT-4 across all five dimensions (Overall: 0.57 vs.\ 0.50), demonstrating that the pedagogical capabilities learned from our training data transfer effectively to an independent evaluation scenario.
Second, compared to SocraticLM~\cite{liu2025socraticlm}, which is specifically trained on SocraTeach data, EduDial-LLM achieves competitive performance under a zero-shot setting (Overall: 0.57 vs.\ 0.62), with the remaining gap primarily attributable to SocraticLM's in-domain training advantage.
Third, the backbone model QwQ-32B-Preview scores substantially lower than GPT-4 (Overall: 0.31 vs.\ 0.50), confirming that the observed performance gains originate from the EduDial training pipeline rather than from the backbone's inherent capabilities.

\subsubsection{Training Corpus Quality Validation}
To validate the effectiveness of our training corpora, we conduct comprehensive ablation experiments examining SFT data quality, DPO data quality, and cross-dataset robustness, as shown in Table~\ref{tab:corpus_quality}.

\begin{table}[!ht]
\centering
\resizebox{\columnwidth}{!}{%
\renewcommand{\arraystretch}{1.0}
\begin{tabular}{l|ccccccccc|c}
\toprule
\rowcolor{gray!15}
\textbf{Model} & \textbf{INS} & \textbf{RES} & \textbf{FB} & \textbf{THK} & \textbf{INT} & \textbf{EMO} & \textbf{ADP} & \textbf{FLU} & \textbf{GOL} & \textbf{AVG} \\
\midrule
\multicolumn{11}{l}{\cellcolor{gray!15}\textit{Base Model}} \\
\midrule
QwQ-32B-Preview & 2.89 & 3.24 & 2.65 & 2.43 & 2.69 & 2.67 & 2.16 & 3.15 & 2.89 & 2.75 \\
\midrule
\multicolumn{11}{l}{\cellcolor{gray!15}\textit{Base Model + SFT}} \\
\midrule
\rowcolor{gray!5}
w/ SocraticLM & 3.64 & 3.71 & 3.68 & 3.56 & 3.63 & 3.58 & 3.65 & 3.98 & 3.76 & 3.69 \\
w/ MTI & 4.12 & 4.35 & 4.28 & 3.88 & 4.14 & 3.90 & 4.13 & 4.79 & 4.49 & 4.23 \\
\midrule
\multicolumn{11}{l}{\cellcolor{gray!15}\textit{Base Model + SFT + DPO (PDTS-Reversed)}} \\
\midrule
\rowcolor{gray!5}
w/ SocraticLM w/ PDTS-Reversed & 3.71 & 3.78 & 3.75 & 3.62 & 3.70 & 3.65 & 3.72 & 4.05 & 3.83 & 3.76 \\
w/ MTI w/ PDTS-Reversed & 4.07 & 4.30 & 4.23 & 3.83 & 4.09 & 3.85 & 4.08 & 4.74 & 4.44 & 4.18 \\
\midrule
\multicolumn{11}{l}{\cellcolor{gray!15}\textit{Base Model + SFT + DPO (PDTS)}} \\
\midrule
\rowcolor{gray!5}
w/ SocraticLM w/ PDTS & 3.95 & 4.02 & 4.00 & 3.87 & 3.94 & 3.89 & 3.96 & 4.29 & 4.07 & 4.00 \\
EduDial-LLM (Ours) & \textbf{4.58} & \textbf{4.12} & \textbf{4.62} & \textbf{4.55} & \textbf{4.64} & \textbf{4.10} & \textbf{4.60} & \textbf{4.43} & \textbf{4.56} & \textbf{4.47} \\
\bottomrule
\end{tabular}
}
\caption{Training Corpus Quality Validation Results. Models are trained on QwQ-32B-Preview with different SFT datasets (SocraticLM or MTI) and DPO datasets (PDTS or PDTS-Reversed: swaps the chosen and rejected responses in PDTS).}
\label{tab:corpus_quality}
\end{table}

As presented in Table~\ref{tab:corpus_quality}, the results demonstrate three key findings. 
First, regarding SFT data quality: our MTI dataset (AVG: 4.23) substantially outperforms SocraticLM (AVG: 3.69) by 0.54 points, representing a 14.6\% improvement over the base model (AVG: 2.75), which confirms MTI's superiority in capturing effective pedagogical interactions.
Second, for DPO data quality: the reversed preference pairs (PDTS-Reversed, AVG: 4.18) underperform the SFT-only baseline (AVG: 4.23), while our PDTS achieves the highest score (AVG: 4.47), validating our preference data construction strategy.
Third, cross-dataset robustness: applying PDTS to SocraticLM-trained models yields 8.4\% improvement (from 3.69 to 4.00) versus only 1.9\% with PDTS-Reversed (from 3.69 to 3.76), demonstrating that our preference dataset generalizes effectively across different models.

\subsubsection{Backbone Model Selection}

To validate our choice of QwQ-32B-Preview as the backbone and demonstrate the generalizability of our training data, we conduct experiments across multiple architectures with varying scales and capabilities, as shown in Table~\ref{tab:backbone_selection}.

\begin{table}[!ht]
\centering
\resizebox{\columnwidth}{!}{%
\renewcommand{\arraystretch}{1.0}
\begin{tabular}{l|ccccccccc|c}
\toprule
\rowcolor{gray!15}
\textbf{Model} & \textbf{INS} & \textbf{RES} & \textbf{FB} & \textbf{THK} & \textbf{INT} & \textbf{EMO} & \textbf{ADP} & \textbf{FLU} & \textbf{GOL} & \textbf{AVG} \\
\midrule
Llama-3.3-70B & 2.56 & 3.08 & 2.53 & 2.36 & 2.81 & 2.72 & 2.47 & 4.25 & 3.56 & 2.93 \\
\rowcolor{gray!5}
Llama-3.3-70B w/ SFT w/ DPO & 3.83 & 4.29 & 3.96 & 4.10 & 4.02 & 4.12 & 4.24 & 4.64 & 4.58 & 4.15 \\
\midrule
DeepSeek-32B & 2.06 & 2.08 & 1.26 & 2.08 & 1.58 & 1.44 & 1.49 & 3.50 & 3.02 & 2.06 \\
\rowcolor{gray!5}
DeepSeek-32B w/ SFT w/ DPO & 3.93 & 3.92 & 3.81 & 4.00 & 3.98 & 4.01 & 4.02 & 4.47 & 4.50 & 4.07 \\
\midrule
Qwen2.5-72B & 2.89 & 2.74 & 2.43 & 2.89 & 2.78 & 2.67 & 2.41 & 3.62 & 3.39 & 2.87 \\
\rowcolor{gray!5}
Qwen2.5-72B w/ SFT w/ DPO & 4.32 & 4.49 & 4.37 & 4.41 & 4.54 & 4.13 & 4.61 & 4.48 & 4.59 & 4.44 \\
\midrule
Qwen2.5-Math & 2.02 & 2.32 & 1.96 & 1.93 & 1.91 & 1.89 & 1.84 & 2.84 & 2.70 & 2.16 \\
\rowcolor{gray!5}
Qwen2.5-Math w/ SFT w/ DPO & 4.58 & 4.08 & 4.49 & 4.22 & 4.58 & 4.08 & 4.38 & 4.24 & 4.27 & 4.32 \\
\midrule
QwQ-32B-Preview & 2.89 & 3.24 & 2.65 & 2.43 & 2.69 & 2.67 & 2.16 & 3.15 & 2.89 & 2.75 \\
\rowcolor{blue!10}
\textbf{EduDial-LLM (Ours)} & \textbf{4.58} & \textbf{4.12} & \textbf{4.62} & \textbf{4.55} & \textbf{4.64} & \textbf{4.10} & \textbf{4.60} & \textbf{4.43} & \textbf{4.56} & \textbf{4.47} \\
\bottomrule
\end{tabular}
}
\caption{Backbone Model Selection Results. Note: ``w/ SFT w/ DPO'' denotes models trained with our MTI dataset for SFT and PDTS dataset for DPO. }
\label{tab:backbone_selection}
\end{table}

As shown in Table~\ref{tab:backbone_selection}, the results validate two key aspects of our model development.
First, QwQ-32B-Preview achieves the best overall performance (AVG: 4.47) after training, justifying its selection as our backbone.
While Qwen2.5-72B achieves competitive performance (AVG: 4.44), QwQ-32B-Preview shows superior balance across pedagogical dimensions, particularly in thinking (4.55 vs. 4.41), interactivity (4.64 vs. 4.54), and adaptability (4.60 vs. 4.61).
Second, the consistent improvements across all backbones after training demonstrate the strong generalizability of our datasets.
The average improvements range from 1.22 (Llama-3.3-70B) to 2.16 (Qwen2.5-Math) points, with DeepSeek-32B showing the most dramatic improvement (2.01 points, 97.6\% relative gain), confirming that our methodology transfers effectively across different architectures and scales.

\subsubsection{Out-of-Distribution Generalization}
To assess our model's robustness with student types not seen during training, we evaluate EduDial-LLM with six student profiles based on SocraticLM's cognitive dimensions, as shown in Table~\ref{tab:ood_generalization}.

\begin{table}[!ht]
\centering
\resizebox{\columnwidth}{!}{%
\renewcommand{\arraystretch}{1.0}
\begin{tabular}{l|ccccccccc|c}
\toprule
\rowcolor{gray!15}
\textbf{Student Type} & \textbf{INS} & \textbf{RES} & \textbf{FB} & \textbf{THK} & \textbf{INT} & \textbf{EMO} & \textbf{ADP} & \textbf{FLU} & \textbf{GOL} & \textbf{AVG} \\
\midrule
Type 1: Poor Problem Understanding & 4.15 & 4.38 & 4.52 & 4.19 & 4.22 & 4.08 & 4.65 & 4.61 & 4.73 & 4.28 \\
\rowcolor{gray!5}
Type 2: Poor Instruction Understanding & 4.21 & 4.42 & 4.48 & 4.16 & 4.27 & 4.11 & 4.58 & 4.92 & 4.79 & 4.38 \\
Type 3: Poor Calculation & 4.19 & 4.45 & 4.56 & 4.22 & 4.31 & 4.15 & 4.71 & 4.68 & 4.81 & 4.36 \\
\rowcolor{gray!5}
Type 4: Poor Knowledge Mastery & 4.12 & 4.47 & 4.49 & 4.14 & 4.19 & 4.06 & 4.73 & 4.65 & 4.76 & 4.29 \\
Type 5: Poor Thirst for Learning & 4.18 & 4.41 & 4.54 & 4.11 & 4.26 & 4.17 & 4.62 & 4.72 & 4.78 & 4.31 \\
\rowcolor{gray!5}
Type 6: Excellent at All Dimensions & 4.25 & 4.46 & 4.63 & 4.28 & 4.38 & 4.14 & 4.69 & 4.74 & 4.87 & 4.42 \\
\midrule
\textbf{Average} & \textbf{4.18} & \textbf{4.43} & \textbf{4.54} & \textbf{4.18} & \textbf{4.27} & \textbf{4.12} & \textbf{4.66} & \textbf{4.72} & \textbf{4.79} & \textbf{4.36} \\
\bottomrule
\end{tabular}
}
\caption{Out-of-Distribution Generalization Results.}
\label{tab:ood_generalization}
\end{table}

As presented in Table~\ref{tab:ood_generalization}, the results demonstrate strong out-of-distribution generalization capability.
EduDial-LLM maintains high teaching quality when interacting with unseen student profiles, achieving an average score of 4.36, only 0.12 points (2.7\% relative decrease) lower than in-distribution students (AVG: 4.48 from Table~\ref{tab:ablation_results}).
Notably, the model exhibits consistent performance across diverse student profiles, with scores ranging from 4.28 to 4.42.
This robustness indicates that our training methodology successfully captures generalizable pedagogical strategies, validating the model's practical applicability to real-world educational scenarios with diverse learner profiles.

\subsubsection{Simulated Student Quality Verification}

To ensure the authenticity of our simulated student behaviors, we conduct expert evaluation assessing the alignment between simulated responses and predefined student profiles across five cognitive dimensions, as shown in Table~\ref{tab:student_simulation_quality}.

\begin{table}[!ht]
\centering
\resizebox{\columnwidth}{!}{%
\renewcommand{\arraystretch}{1.0}
\begin{tabular}{l|ccccc|c}
\toprule
\rowcolor{gray!15}
\textbf{Student Profile} & \textbf{Cognitive} & \textbf{Motivation} & \textbf{Participation} & \textbf{Strategies} & \textbf{Self-Reg.} & \textbf{Average} \\
\midrule
Excellent Student & 4.58 & 4.76 & 4.68 & 4.52 & 4.73 & 4.65 \\
\rowcolor{gray!5}
Medium Student & 4.43 & 4.62 & 3.96 & 4.48 & 4.58 & 4.41 \\
Struggling Student & 4.47 & 4.71 & 3.82 & 4.45 & 4.76 & 4.44 \\
\midrule
\textbf{Average} & \textbf{4.49} & \textbf{4.69} & \textbf{4.15} & \textbf{4.48} & \textbf{4.69} & \textbf{4.48} \\
\bottomrule
\end{tabular}
}
\caption{Simulated Student Quality Verification Results.}
\label{tab:student_simulation_quality}
\end{table}

As shown in Table~\ref{tab:student_simulation_quality}, the expert evaluation results validate the strong alignment between simulated behaviors and predefined student profiles.
All three student types achieve scores ranging from 3.82 to 4.76 across five cognitive dimensions, with an overall average of 4.48, indicating strong alignment between simulated behaviors and predefined profiles.
The inter-rater reliability measured by Cohen's Kappa coefficient is 0.71, demonstrating substantial agreement among expert evaluations.
Notably, learning motivation and self-regulation dimensions receive the highest ratings (both AVG: 4.69), suggesting our simulation effectively captures students' internal learning drives and metacognitive characteristics.
Classroom participation shows relatively lower scores (AVG: 4.15), particularly for struggling students (3.82), which authentically reflects the typical behavioral patterns of learners with different proficiency levels.
These results confirm that our simulated students provide realistic and reliable training data for developing effective teaching models.

\subsubsection{Balancing Teaching Capability and Mathematical Reasoning}

To verify our model maintains general reasoning capability while enhancing teaching effectiveness, we evaluate performance on challenging mathematical reasoning benchmarks, as shown in Table~\ref{tab:math_benchmarks}.

\begin{table}[t]
\centering
\resizebox{\columnwidth}{!}{%
\renewcommand{\arraystretch}{1.0}
\begin{tabular}{l|c|c}
\toprule
\rowcolor{gray!15}
\textbf{Model} & \textbf{Math500} & \textbf{AIME2024} \\
\midrule
QwQ-32B-Preview & \textbf{91.12} & \textbf{43.33} \\
\rowcolor{gray!5}
EduDial-LLM & 87.54 & 33.33 \\ 
\bottomrule
\end{tabular}
}
\caption{Reasoning Capability of Base Model and EduDial-LLM on Standard Benchmarks.}
\label{tab:math_benchmarks}
\end{table}

As presented in Table~\ref{tab:math_benchmarks}, EduDial-LLM maintains acceptable mathematical reasoning capabilities while prioritizing pedagogical effectiveness.
Our model achieves 87.54\% on Math500 (3.58 percentage points below baseline) and 33.33\% on AIME2024 (10 percentage points below baseline).
Although performance decreases on pure reasoning tasks, this trade-off is expected and justified given our optimization objective.
In educational contexts, an effective teaching assistant requires both accurate problem-solving abilities and pedagogical skills such as identifying confusion points and providing personalized guidance.
EduDial-LLM maintains competitive reasoning performance while excelling in teaching-specific dimensions (AVG: 4.47 vs. baseline: 2.75), making it more suitable for real-world educational applications.

\subsubsection{Case Study}

To illustrate the effectiveness of our approach, we present a comparative case study examining how EduDial-LLM and GPT-4o guide students in learning the concept of "Normal Distribution."

\begin{tcolorbox}[
    colback=white,
    colframe=gray!60!black,
    title={\textbf{EduDial-LLM 32B: Teaching Normal Distribution}},
    fonttitle=\bfseries\small,
    coltitle=black,
    boxrule=1pt,
    arc=2mm,
    left=2mm,
    right=2mm,
    top=2mm,
    bottom=2mm,
    breakable=true
]
\small
\textbf{Teacher:} Students, today we're going to learn about a new statistical concept — normal distribution. Have you observed any patterns in the distribution of phenomena or data? For example, how do you think the test scores in your class might be distributed?

\vspace{0.3em}
\textbf{Excellent Student:} I think there would probably be an average value, with most students' scores clustered around it, and only a few doing exceptionally well or poorly.

\vspace{0.3em}
\textbf{Medium Student:} I think test scores would mostly be concentrated in the middle position, with possibly only a few students scoring particularly high or low.

\vspace{0.3em}
\textbf{Struggling Student:} I'm not really sure, but I feel like some students would do well and some would do poorly.

\vspace{0.3em}
\textbf{Teacher:} Very good! Your observations are very close to the normal distribution we're going to learn about today. Normal distribution is commonly used to describe many real-life phenomena, such as test scores, human height, weight, etc. Its characteristic is that most data is concentrated around the mean value, gradually decreasing on both sides. In our previous studies, we've already encountered the concepts of random variables and probability distributions. Do you remember what a probability distribution is?
\end{tcolorbox}

\vspace{0.5em}

\begin{tcolorbox}[
    colback=white,
    colframe=gray!60!black,
    title={\textbf{GPT-4o: Teaching Normal Distribution}},
    fonttitle=\bfseries\small,
    coltitle=black,
    boxrule=1pt,
    arc=2mm,
    left=2mm,
    right=2mm,
    top=2mm,
    bottom=2mm,
    breakable=true
]
\small
\textbf{Teacher:} Today, we are learning about normal distribution. Has anyone seen a "high in the middle, low on both sides" bell-shaped distribution in daily life?

\vspace{0.3em}
\textbf{Excellent Student:} Yes! Test scores, long jump results, and height data all follow this pattern, with most people clustered around the average value.

\vspace{0.3em}
\textbf{Medium Student:} Yes, class heights and test scores indeed follow this pattern, with fewer students getting very high or very low scores.

\vspace{0.3em}
\textbf{Struggling Student:} I think I've heard that test scores are like this... But what does this have to do with normal distribution? Is it very complicated?

\vspace{0.3em}
\textbf{Teacher:} Normal distribution can be represented by a bell-shaped curve, and its formula is: 

\vspace{0.3em}
\begin{equation*}
\phi_{\mu,\sigma}(x) = \frac{1}{\sigma\sqrt{2\pi}}e^{-\frac{(x-\mu)^2}{2\sigma^2}}
\end{equation*}

where $\mu$ is the center position (mean), and $\sigma$ is the width of the distribution (standard deviation).
\end{tcolorbox}

This comparison reveals significant pedagogical differences between the two approaches. EduDial-LLM employs an open-ended inquiry strategy that activates students' prior knowledge and encourages observation-based learning, progressively building from students' existing understanding to formal mathematical concepts. In contrast, GPT-4o follows a more traditional didactic approach, directly introducing the concept with pre-defined examples before immediately presenting the complex mathematical formula. Notably, EduDial-LLM demonstrates adaptive teaching by acknowledging all student responses and building upon each student's understanding level, while GPT-4o maintains a uniform teaching pace regardless of individual cognitive differences. Through this case study, we observe that EduDial-LLM's approach encourages deeper reflection and student engagement through exploratory questions, adapting its teaching strategy to provide appropriate scaffolding for different learner types, thereby embodying the differentiated instruction principles that are fundamental to effective mathematics education.

\subsection{Pedagogical Framework}
\begin{table}[!ht]
\centering
\resizebox{\columnwidth}{!}{%
\renewcommand{\arraystretch}{1.1}
\begin{tabular}{p{3.5cm}|p{4.5cm}|p{4cm}|p{7cm}}
\toprule
\rowcolor{gray!15} 
\textbf{Question Type} & \textbf{Purpose} & \textbf{Cognitive Process} & \textbf{Exemplar Questions} \\
\midrule
\textbf{Scenario-based} 
& Connect abstract concepts to real-world applications 
& Activation of prior knowledge → Contextualization 
& "In planning a 'Community Garden Project' with varied participant skills (vegetable growing, design, tool provision), how would you optimize team allocation?" \\
\midrule
\rowcolor{gray!5}
\textbf{Retrospective} 
& Bridge previous and new knowledge 
& Memory retrieval → Knowledge integration 
& "How does today's set theory connect with our earlier classification of numbers?" \\
\midrule
\textbf{Predictive} 
& Stimulate hypothesis formation 
& Pattern recognition → Logical conjecture 
& "Predict the elements when grouping all even numbers from 1 to 10 into a set" \\
\midrule
\rowcolor{gray!5}
\textbf{Verification} 
& Develop critical evaluation skills 
& Hypothesis testing → Counterexample analysis 
& "If we add 0 to the positive integer set, does it remain a positive integer set? Justify." \\
\midrule
\textbf{Zone of Proximal Development} 
& Scaffold conceptual advancement 
& Guided discovery → Generalization 
& "Extend the 1-10 even number set to formulate a general representation for any range" \\
\midrule
\rowcolor{gray!5}
\textbf{Critical} 
& Foster alternative perspectives 
& Divergent thinking → Justification 
& "Design a novel integer classification beyond odd/even with explanatory principles" \\
\midrule
\textbf{Integrative} 
& Synthesize multiple concepts 
& Cross-domain application → System thinking 
& "Find a number in $\mathbb{Q}$ but not $\mathbb{Z}$ and explain its significance" \\
\midrule
\rowcolor{gray!5}
\textbf{Open-ended} 
& Encourage creative exploration 
& Speculative reasoning → Perspective-taking 
& "How would a mathematical system with only $\mathbb{Z}$ and irrationals impact real-world applications?" \\
\midrule
\textbf{Summative} 
& Consolidate learning outcomes 
& Knowledge organization → Key point identification 
& "Summarize the core concepts of set representation covered today" \\
\midrule
\rowcolor{gray!5}
\textbf{Metacognitive} 
& Develop self-regulated learning 
& Process reflection → Strategy planning 
& "Which aspects do you feel confident about? What improvement plan would address gaps?" \\
\bottomrule
\end{tabular}
}
\caption{Pedagogical Questioning Strategies for the Five-Stage Teaching Process.}
\label{tab:questioning_strategies}
\end{table}

\begin{table}[!ht]
\centering
\resizebox{\columnwidth}{!}{%
\renewcommand{\arraystretch}{1.1}
\begin{tabular}{p{3cm}|p{3.5cm}|p{3.5cm}|p{3.5cm}|p{3.5cm}}
\toprule
\rowcolor{gray!15} \textbf{Dimensions} & \textbf{Teacher} & \textbf{Excellent Student} & \textbf{Medium Student} & \textbf{Struggling Student} \\
\midrule
\textbf{Cognitive Ability} & Rigorous understanding of mathematical concepts; capable of clear and accurate explanations & Sharp logical thinking; quickly comprehends new content; applies knowledge to solve complex problems & Good comprehension but requires more time to absorb new knowledge; capable of understanding core content & Weak foundational knowledge; difficulties grasping new concepts; struggles with complex problem-solving \\
\midrule
\rowcolor{gray!5}
\textbf{Learning Motivation} & Committed to comprehensive knowledge transfer; provides emotional support to enhance student confidence & Strong curiosity for knowledge; explores mathematics beyond textbooks; seeks interdisciplinary connections & Moderate interest in learning; requires occasional encouragement to maintain engagement & Lacks interest in mathematics; easily frustrated when unable to keep pace; diminished learning drive \\
\midrule
\textbf{Classroom Participation} & Facilitates active discussions; creates an inclusive learning environment; provides real-time feedback & Active participant in discussions; assumes leadership roles in group learning; helps peers understand difficult concepts & Moderate participation level; contributes when prompted; neither overly active nor passive & Easily distracted; minimal voluntary participation; requires significant prompting to engage \\
\midrule
\rowcolor{gray!5}
\textbf{Learning Strategies} & Employs diverse teaching methodologies; adjusts explanations to accommodate different learning levels; uses varied assessment techniques & Self-regulated learning; effectively balances academic and extracurricular activities; employs multiple problem-solving approaches & Limited range of learning strategies; relies on note-taking and peer discussions; benefits from structured learning frameworks & Highly dependent on external support; lacks independent problem-solving skills; requires step-by-step guidance \\
\midrule
\textbf{Self-Regulation} & Monitors classroom dynamics; adapts teaching pace and content based on student responses; balances coverage with comprehension & Self-disciplined; maintains consistent learning motivation despite challenges; sets clear goals with actionable implementation plans & Requires external structure; occasionally lacks confidence, especially with complex problems; benefits from incremental successes & Strong dependency on external motivation; easily discouraged by failures; shows improvement potential with personalized guidance and encouragement \\
\bottomrule
\end{tabular}
}
\caption{Role Profiles for Teaching Scenarios.}
\label{tab:role_profiles}
\end{table}

\begin{table}[!ht]
\centering
\resizebox{\columnwidth}{!}{%
\renewcommand{\arraystretch}{1.1}
\begin{tabular}{p{8cm}|p{8cm}}
\toprule
\rowcolor{gray!15} \textbf{Student Behavior} & \textbf{Teacher Strategy} \\
\midrule
\textbf{Excellent Students} & \textbf{Depth Exploration Strategy} \\
\midrule
Inquiry Questions & Guide students to understand the concept essence \\
\rowcolor{gray!5}
Extension Questions & Guide independent research and thinking \\
Optimization Questions & Compare solutions from multiple perspectives \\
\midrule
Accurate answers with explanations & Design advanced questions for deeper understanding \\
\rowcolor{gray!5}
Exploring essence & Explore core elements and mechanisms \\
Connecting relevant knowledge & Establish connections across contexts \\
\midrule
\textbf{Medium Students} & \textbf{Guided Thinking Strategy} \\
\midrule
Application Questions & Help understand knowledge connections \\
\rowcolor{gray!5}
Technique Questions & Explain logic and techniques of problem-solving \\
Detail Questions & Analyze inquiries to consolidate knowledge \\
\midrule
Partially correct answers & Affirm correct portions, provide feedback \\
\rowcolor{gray!5}
Expressing uncertainty & Break complex problems into sub-steps \\
Seeking hints & Guide analysis of key components \\
\midrule
\textbf{Struggling Students} & \textbf{Confidence Building Strategy} \\
\midrule
Basic Questions & Emphasize fundamentals with simple methods \\
\rowcolor{gray!5}
Comprehension Questions & Guide problem analysis for understanding \\
Learning Efficacy Questions & Use examples to build confidence \\
\midrule
Incorrect answers & Provide encouragement and gradual guidance \\
\rowcolor{gray!5}
Vague answers & Use prompts to clarify reasoning \\
Unable to answer & Begin with basics to reduce pressure \\
\bottomrule
\end{tabular}
}
\caption{Teaching Strategies for Different Student Question and Reply Types.}
\label{tab:teaching_strategies}
\end{table}

\subsection{Prompt Templates}

\begin{tcolorbox}[title=Generation MTI Dataset Prompt Template, fontupper=\small, breakable=true]

\section*{Student Profiles}
\begin{itemize}[leftmargin=*]
    \item \textbf{Excellent Student (s1):} \{Excellent profile\}
    \item \textbf{Medium Student (s2):} \{Medium student profile\}
    \item \textbf{Struggling Student (s3):} \{Struggling student profile\}
\end{itemize}

\section*{Teacher Profile}
\begin{itemize}[leftmargin=*]
    \item \textbf{Teacher Profile:} \{Teacher Profile\}
\end{itemize}

\section*{Teaching Stages \& Questioning Strategies}
\begin{enumerate}[leftmargin=*]
    \item \textbf{Introduction Stage}
    \begin{itemize}
        \item \textit{Scenario-based Questions:} Guide students to understand concepts through life experiences\\
        Example: \{Scenario question example\}
        \item \textit{Recall Questions:} Help students connect prior knowledge with new concepts\\
        Example: \{Recall question example\}
    \end{itemize}
    
    \item \textbf{Concept Exploration Stage}
    \begin{itemize}
        \item \textit{Prediction Questions:} Encourage students to predict new concepts based on existing knowledge\\
        Example: \{Prediction question example\}
        \item \textit{Verification Questions:} Verify conceptual understanding through examples\\
        Example: \{Verification question example\}
    \end{itemize}
    
    \item \textbf{Deep Understanding Stage}
    \begin{itemize}
        \item \textit{Zone of Proximal Development Questions:} Present questions slightly above students' current ability\\
        Example: \{ZPD question example\}
        \item \textit{Critical Thinking Questions:} Encourage students to question assumptions and explore new perspectives\\
        Example: \{Critical thinking question example\}
    \end{itemize}
    
    \item \textbf{Knowledge Application Stage}
    \begin{itemize}
        \item \textit{Integrative Questions:} Combine multiple knowledge points to solve complex problems\\
        Example: \{Integrative question example\}
        \item \textit{Open-ended Questions:} Present questions without unique answers to promote creative thinking\\
        Example: \{Open-ended question example\}
    \end{itemize}
    
    \item \textbf{Reflection and Summary Stage}
    \begin{itemize}
        \item \textit{Summary Questions:} Help students review and consolidate understanding\\
        Example: \{Summary question example\}
        \item \textit{Metacognitive Questions:} Guide students to evaluate their learning methods and effectiveness\\
        Example: \{Metacognitive question example\}
    \end{itemize}
\end{enumerate}

\section*{Teaching Syllabus}
\{Detailed teaching syllabus description\}

\section*{Interaction Structure}
\begin{itemize}[leftmargin=*]
    \item At least two rounds of dialogue per phase
    \item Ensure coherence between phases
\end{itemize}

\end{tcolorbox}

\begin{tcolorbox}[title=Deep Exploration Strategy Prompt Template for Excellent Students, fontupper=\small, breakable=true]

You are a mathematics teacher delivering a lesson based on the curriculum: \{Teaching Content\}

The classroom instruction is divided into five phases, with teacher (T) and student interactions in each phase:

\begin{itemize}
    \item \textbf{T}: Represents the teacher, possessing systematic mathematical knowledge, rich teaching experience, and differentiated teaching abilities.
    \item \textbf{s1}: Represents high-performing students, typically with active thinking, solid foundation, and questions involving deep thinking or knowledge extension.
    \item \textbf{s2}: Represents average students, with relatively stable basic knowledge, but possibly insufficient mastery of problem-solving techniques and comprehensive application abilities.
    \item \textbf{s3}: Represents struggling students, who may have difficulty keeping up with the classroom pace due to weak foundations or comprehension deficiencies.
\end{itemize}

\section*{High-Performing Student (s1) Characteristics and Teaching Strategies}

\subsection*{Student Performance Characteristics:}

\begin{enumerate}
    \item \textbf{Question Types:}
    \begin{itemize}
        \item \textbf{Inquiry questions}: e.g., "Why can derivatives represent the rate of change of a function? How is this conclusion derived?"
        \item \textbf{Extension questions}: e.g., "Can this theorem be applied in more complex situations, such as non-linear equations?"
        \item \textbf{Optimization questions}: e.g., "Is the standard answer to this problem unique? Is there a more concise solution?"
    \end{itemize}
    
    \item \textbf{Response Types:}
    \begin{itemize}
        \item \textbf{Accurate answers with explanations}: e.g., "This function has an extreme point at x = 1 because the derivative is zero at that point and the second derivative is greater than zero, so it's a minimum value."
        \item \textbf{Exploring conceptual essence}: e.g., "Besides the algebraic method, we can also use a geometric method to prove this conclusion."
        \item \textbf{Creating knowledge connections}: e.g., "This problem is related to the arithmetic sequence we learned earlier, and can be solved using the general term formula for arithmetic sequences."
    \end{itemize}
\end{enumerate}

\subsection*{Teacher Response Strategy (Deep Exploration Strategy):}

\begin{itemize}
    \item Guide students to consider theoretical foundations, helping them understand the origins and deeper meanings of concepts
    \item Provide extended explanations or guide independent research, cultivating independent thinking abilities
    \item Encourage reflection on problems from different angles, exploring multiple solutions, fostering critical and innovative thinking
    \item Design higher-order challenging questions, encouraging application of learned knowledge to solve open-ended or comprehensive problems
    \item Guide students to reflect on problem-solving processes, promoting group discussion and collaboration
\end{itemize}

\section*{Task Requirements}

Design 25 non-repetitive single-round dialogues for high-performing students (s1) across five classroom phases, with 5 dialogues per phase. Each dialogue should include:

\begin{enumerate}
    \item s1's statement (question or answer): Must conform to high-performing student characteristics, reflecting realistic teaching scenarios, limited to mathematical content
    \item Teacher's response (T): Apply the deep exploration strategy, demonstrate teaching professionalism, and provide detailed guidance for student thinking processes
\end{enumerate}

Generated dialogue format should be [s1], [T], ensuring rich, deep content that reflects the teacher's guidance for high-performing students' cognitive development.

\end{tcolorbox}

\begin{tcolorbox}[title=Thinking Guidance Strategy Prompt Template for Average Students, fontupper=\small, breakable=true]

You are a mathematics teacher delivering a lesson based on the curriculum: \{Teaching Content\}

The classroom instruction is divided into five phases, with teacher (T) and student interactions in each phase:

\begin{itemize}
    \item \textbf{T}: Represents the teacher, possessing systematic mathematical knowledge, rich teaching experience, and differentiated teaching abilities.
    \item \textbf{s1}: Represents high-performing students, typically with active thinking, solid foundation, and questions involving deep thinking or knowledge extension.
    \item \textbf{s2}: Represents average students, with relatively stable basic knowledge, but possibly insufficient mastery of problem-solving techniques and comprehensive application abilities.
    \item \textbf{s3}: Represents struggling students, who may have difficulty keeping up with the classroom pace due to weak foundations or comprehension deficiencies.
\end{itemize}

\section*{Average Student (s2) Characteristics and Teaching Strategies}

\subsection*{Student Performance Characteristics:}

\begin{enumerate}
    \item \textbf{Question Types:}
    \begin{itemize}
        \item \textbf{Application questions}: e.g., "In this problem, how do I determine whether to use the binomial theorem or direct expansion?"
        \item \textbf{Skill-based questions}: e.g., "Why must this problem be simplified first rather than directly substituting into the formula?"
        \item \textbf{Detail-oriented questions}: e.g., "Why can this transformation step be done this way?"
    \end{itemize}
    
    \item \textbf{Response Types:}
    \begin{itemize}
        \item \textbf{Partially correct answers}: e.g., "For the equation x²-4=0, after moving terms and taking the square root, we get x=2."
        \item \textbf{Expressing uncertainty}: e.g., "I think I should use the formula to solve it, but I don't know how to proceed to the next step."
        \item \textbf{Seeking hints}: e.g., "This problem is a bit difficult, could you give me another hint?"
    \end{itemize}
\end{enumerate}

\subsection*{Teacher Response Strategy (Thinking Guidance Strategy):}

\begin{itemize}
    \item Help students understand internal connections between knowledge points, clarifying their functions and applicable scenarios
    \item Gradually explain the logic of problem-solving processes, encourage independent restatement of problem-solving ideas, cultivating thinking abilities
    \item Provide in-depth explanations for detail-oriented questions, offer efficient practice strategies to help consolidate knowledge application
    \item Provide positive feedback, promptly praise progress and correct portions, offering improvement suggestions in an encouraging manner
    \item Provide layered guidance, breaking down problem-solving processes into smaller steps, guiding completion step-by-step
\end{itemize}

\section*{Task Requirements}

Design 25 non-repetitive single-round dialogues for average students (s2) across five classroom phases, with 5 dialogues per phase. Each dialogue should include:

\begin{enumerate}
    \item s2's statement (question or answer): Must conform to average student characteristics, reflecting realistic teaching scenarios, limited to mathematical content
    \item Teacher's response (T): Apply the thinking guidance strategy, demonstrate teaching professionalism, and provide detailed guidance for problem-solving approaches
\end{enumerate}

Generated dialogue format should be [s2], [T], ensuring responses are inspirational and effectively help average students understand knowledge points and improve problem-solving abilities.

\end{tcolorbox}

\begin{tcolorbox}[title=Confidence Building Strategy Prompt Template for Struggling Students, fontupper=\small, breakable=true]

You are a mathematics teacher delivering a lesson based on the curriculum: \{Teaching Content\}

The classroom instruction is divided into five phases, with teacher (T) and student interactions in each phase:

\begin{itemize}
    \item \textbf{T}: Represents the teacher, possessing systematic mathematical knowledge, rich teaching experience, and differentiated teaching abilities.
    \item \textbf{s1}: Represents high-performing students, typically with active thinking, solid foundation, and questions involving deep thinking or knowledge extension.
    \item \textbf{s2}: Represents average students, with relatively stable basic knowledge, but possibly insufficient mastery of problem-solving techniques and comprehensive application abilities.
    \item \textbf{s3}: Represents struggling students, who may have difficulty keeping up with the classroom pace due to weak foundations or comprehension deficiencies.
\end{itemize}

\section*{Struggling Student (s3) Characteristics and Teaching Strategies}

\subsection*{Student Performance Characteristics:}

\begin{enumerate}
    \item \textbf{Question Types:}
    \begin{itemize}
        \item \textbf{Fundamental questions}: e.g., "Why are alternate interior angles equal when two parallel lines are crossed by a transversal?"
        \item \textbf{Understanding questions}: e.g., "I'm completely lost, where should I start with this problem?"
    \end{itemize}
    
    \item \textbf{Response Types:}
    \begin{itemize}
        \item \textbf{Incorrect answers}: e.g., "Teacher, the expansion of (a+b)² is 2a+2b."
        \item \textbf{Vague responses}: e.g., "I think I calculated something wrong, I'm not sure if the answer is correct."
        \item \textbf{Direct expression of inability}: e.g., "Teacher, I don't know how to do this."
    \end{itemize}
\end{enumerate}

\subsection*{Teacher Response Strategy (Confidence Building Strategy):}

\begin{itemize}
    \item Emphasize repeated explanation of basic knowledge, using straightforward language to aid understanding
    \item Patiently guide step-by-step analysis of problems, avoiding presenting too much information at once
    \item Use simple, concrete examples to help build confidence and reduce frustration
    \item Affirm students' efforts and correct portions, then help gradually correct thinking paths, protecting self-confidence
    \item Start from the most basic elements, allowing students to experience step-by-step success, reducing psychological burden
\end{itemize}

\section*{Task Requirements}

Design 25 non-repetitive single-round dialogues for struggling students (s3) across five classroom phases, with 5 dialogues per phase. Each dialogue should include:

\begin{enumerate}
    \item s3's statement (question or answer): Must conform to struggling student characteristics, reflecting realistic teaching scenarios, limited to mathematical content
    \item Teacher's response (T): Apply the confidence-building strategy, demonstrating teaching patience and understanding of student psychology
\end{enumerate}

Generated dialogue format should be [s3], [T], ensuring responses are clear, progressive, and help struggling students build confidence in mathematics learning.

\end{tcolorbox}

\begin{tcolorbox}[title=Evaluation Prompt Template, fontupper=\small, breakable=true, enhanced]

Please evaluate the following teaching content based on 11 teaching indicators spanning two dimensions: overall quality and content quality. Each overall quality indicator and relevance score ranges from 1-5 points (1 being the lowest, 5 being the highest). Coverage is rated from 0\%-100\% (0\% being the lowest, 100\% being the highest).

\section*{Assessment Object}
Teaching content: \{Teaching content\}\\
Teaching outline: \{Teaching outline\}

\section*{Overall Quality Dimension (1-5 points each)}

\subsection*{1. Insight}
\textbf{Definition:} Assessing whether the teacher can accurately capture and deeply understand students' learning needs, knowledge levels, and question intentions.\\
\textbf{Assessment points:}
\begin{itemize}[leftmargin=*]
    \item \textbf{Need identification:} Whether the teacher identifies students' specific learning needs and difficulties through questioning, observation, or other methods
    \item \textbf{Problem analysis:} Whether the teacher can analyze students' questions and understand the learning obstacles or misconceptions behind them
    \item \textbf{Personalized understanding:} Whether the teacher demonstrates understanding of each student's unique learning style and needs
\end{itemize}
\textbf{Score:} [ ]/5

\subsection*{2. Response}
\textbf{Definition:} Assessing whether the teacher can effectively solve students' problems and provide practical and constructive guidance.\\
\textbf{Assessment points:}
\begin{itemize}[leftmargin=*]
    \item \textbf{Problem solving:} Whether the teacher can provide clear and effective solutions to students' problems
    \item \textbf{Specificity of guidance:} Whether the advice provided is specific, actionable, and can help students improve in practice
    \item \textbf{Resource provision:} Whether the teacher recommends relevant resources (such as textbooks, exercises, reference materials) to support students' further learning
\end{itemize}
\textbf{Score:} [ ]/5

\subsection*{3. Feedback}
\textbf{Definition:} Assessing whether the teacher can provide timely and effective feedback to help students improve their learning.\\
\textbf{Assessment points:}
\begin{itemize}[leftmargin=*]
    \item \textbf{Timeliness:} Whether feedback is provided promptly after students raise questions, avoiding delays in students' learning progress
    \item \textbf{Constructiveness:} Whether the feedback content is constructive and can guide students on how to improve
    \item \textbf{Two-way communication:} Whether the teacher encourages students to respond to feedback, promoting two-way communication
\end{itemize}
\textbf{Score:} [ ]/5

\subsection*{4. Thinking}
\textbf{Definition:} Assessing whether the teacher can stimulate students' critical thinking abilities, including analytical ability, open-mindedness, and self-assessment ability.\\
\textbf{Assessment points:}
\begin{itemize}[leftmargin=*]
    \item \textbf{Question guidance:} Whether the teacher guides students to think deeply through open-ended questions
    \item \textbf{Analysis training:} Whether the teacher cultivates students' analytical abilities, helping them break down complex problems
    \item \textbf{Self-assessment:} Whether the teacher encourages students to reflect and self-assess to enhance independent learning abilities
\end{itemize}
\textbf{Score:} [ ]/5

\subsection*{5. Interactivity}
\textbf{Definition:} Assessing the frequency and quality of teacher-student interactions in teaching dialogues, including questioning techniques and ways of responding to student feedback.\\
\textbf{Assessment points:}
\begin{itemize}[leftmargin=*]
    \item \textbf{Questioning techniques:} Whether the teacher uses effective questioning methods to promote student thinking and participation
    \item \textbf{Response to feedback:} Whether the teacher actively responds to student feedback, promoting continued interaction
    \item \textbf{Interaction frequency:} Whether interactions between teacher and students are frequent, avoiding one-way knowledge transmission
\end{itemize}
\textbf{Score:} [ ]/5

\subsection*{6. Emotional Support}
\textbf{Definition:} Assessing whether the teacher can provide emotional support during the teaching process, establish a positive learning environment, and help students build confidence and positive learning attitudes.\\
\textbf{Assessment points:}
\begin{itemize}[leftmargin=*]
    \item \textbf{Emotional care:} Whether the teacher pays attention to students' emotional states and provides necessary care and support
    \item \textbf{Motivation and encouragement:} Whether the teacher stimulates students' learning motivation and self-confidence through encouragement and praise
    \item \textbf{Building trust:} Whether the teacher establishes trust relationships with students, making them feel safe and respected
\end{itemize}
\textbf{Score:} [ ]/5

\subsection*{7. Adaptability}
\textbf{Definition:} Assessing the teacher's ability to adjust teaching methods according to different students' learning styles and needs, ensuring that each student receives personalized guidance.\\
\textbf{Assessment points:}
\begin{itemize}[leftmargin=*]
    \item \textbf{Teaching method adjustment:} Whether the teacher adjusts teaching strategies and methods based on student feedback and learning progress
    \item \textbf{Personalized guidance:} Whether the teacher can provide personalized guidance and support based on different students' characteristics
    \item \textbf{Flexible response:} Whether the teacher can flexibly respond to unexpected situations in the classroom to ensure teaching effectiveness
\end{itemize}
\textbf{Score:} [ ]/5

\subsection*{8. Fluency}
\textbf{Definition:} Assessing whether the teacher's expression is clear, easy to understand, and natural in tone.\\
\textbf{Assessment points:}
\begin{itemize}[leftmargin=*]
    \item \textbf{Language expression:} Whether the language used by the teacher is accurate and concise, avoiding ambiguous or obscure expressions
    \item \textbf{Logical structure:} Whether the teacher's explanation has a clear logical structure that facilitates student understanding
    \item \textbf{Natural tone:} Whether the teacher's tone in communication is friendly and patient, creating a good communication atmosphere
\end{itemize}
\textbf{Score:} [ ]/5

\subsection*{9. Goal}
\textbf{Definition:} Assessing whether the teacher's guidance helps achieve predetermined teaching objectives, such as knowledge mastery, skill development, etc.\\
\textbf{Assessment points:}
\begin{itemize}[leftmargin=*]
    \item \textbf{Clear objectives:} Whether teaching activities have clearly set specific learning objectives
    \item \textbf{Goal alignment:} Whether the teacher's teaching behaviors and guidance align with the set objectives
    \item \textbf{Outcome assessment:} Whether the teacher uses assessment methods to verify if students have achieved the predetermined objectives
\end{itemize}
\textbf{Score:} [ ]/5

\section*{Content Quality Dimension}

\subsection*{10. Relevance}
\textbf{Definition:} Assessing whether the teaching content is relevant to the teaching outline of this lesson.\\
\textbf{Assessment points:}
\begin{itemize}[leftmargin=*]
    \item Whether the content aligns with the themes and objectives of the teaching outline
    \item Whether the content includes core concepts and knowledge points related to the course topic
\end{itemize}
\textbf{Score:} [ ]/5

\subsection*{11. Coverage}
\textbf{Definition:} Assessing the proportion of teaching outline knowledge points covered by the teaching content.\\
\textbf{Assessment points:}
\begin{itemize}[leftmargin=*]
    \item The proportion of knowledge points in the teaching outline covered by the teaching content
    \item Whether the core concepts and key knowledge points in the teaching outline are covered
\end{itemize}
\textbf{Coverage rate:} [ ]\%

\section*{Summary Evaluation}
[Provide comprehensive evaluation and improvement suggestions here]
\end{tcolorbox}

\subsection{The Use of Large Language Models}
In this paper, we utilized LLMs for language polishing to
enhance clarity, and we manually reviewed all modifications.

\subsection{Evaluator Selection and Remuneration}
We recruited the human evaluators through public job postings on local educational forums. All participants are certified mathematics teachers 
with 3-10 years of K-12 teaching experience. The evaluator pool consists 
of 3 females and 2 males, aged between 28 and 45 years old, holding 
Master's degrees in Mathematics. They were compensated at a rate of \$10 per hour. 
The hourly rate exceeds the local minimum wage and is comparable to the average compensation for similar annotation tasks in the region.